\documentclass[letterpaper, 10 pt, journal, twoside]{IEEEtran}
\usepackage{amsmath} 
\usepackage{algorithm}
\usepackage{array}
\usepackage[caption=false,font=normalsize,labelfont=sf,textfont=sf]{subfig}

\usepackage{url}
\usepackage{stfloats}

\ifCLASSINFOpdf
   \usepackage[pdftex]{graphicx}
\else
  \usepackage[dvips]{graphicx}
\fi
\usepackage{cite}
\begin{document}
\title{Trifocal Tensor and Relative Pose Estimation with Known Vertical Direction}
\author{Tao Li$^{1}$, Zhenbao Yu$^{2,3}$, Banglei Guan$^{2*}$, Jianli Han$^{1}$, Weimin Lv$^{1}$ and Friedrich Fraundorfer$^{4}$
\thanks{This paper was recommended for publication by Editor Pascal Vasseur upon evaluation of the Associate Editor and Reviewers' comments.
This work was supported in part by the Hunan Provincial Natural Science Foundation for Excellent Young Scholars under Grant 2023JJ20045, and the National Natural Science Foundation of China under Grant 12372189. 
{Corresponding author: Banglei Guan.}} 
\thanks{$^{1}$Tao Li, Jianli Han, and Weimin Lv are with the College of Aerospace Science and Engineering, Naval Aeronautical University, Yantai 264000, China
{\tt\footnotesize litao0931@alumni.nudt.edu.cn, jianlihan1585@163.com, 2016150315@jou.edu.cn}}
\thanks{$^{2,3} $Zhenbao Yu is with the College of Aerospace Science and Engineering, National University of Defense Technology, Changsha 410000, China and the Global Navigation Satellite System Research Center, Wuhan University, Wuhan 430000, China
{\tt\footnotesize zhenbaoyu@whu.edu.cn}}
\thanks{$^{2} $Banglei Guan is with the College of Aerospace Science and Engineering, National University of Defense Technology, Changsha 410000, China
{\tt\footnotesize guanbanglei12@nudt.edu.cn}}
\thanks{$^{4} $Friedrich Fraundorfer is with the  Institute of Computer Graphics and Vision, Graz University of Technology, Graz, Austria
{\tt\footnotesize friedrich.fraundorfer@tugraz.at}}
}

\markboth{}
{Li \MakeLowercase{\textit{et al.}}: Trifocal Tensor and Relative Pose Estimation with Known Vertical Direction} 
\maketitle
\begin{abstract}
This work presents two novel solvers for estimating the relative poses among views with known vertical directions. The vertical directions of camera views can be easily obtained using inertial measurement units (IMUs) which have been widely used in autonomous vehicles, mobile phones, and unmanned aerial vehicles (UAVs). Given the known vertical directions, our algorithms only need to solve for two rotation angles and two translation vectors.
In this paper, a linear closed-form solution has been described, requiring only four point correspondences in three views. 
We also propose a minimal solution with three point correspondences using the latest Gröbner basis solver. Since the proposed methods require fewer point correspondences, they can be efficiently applied within the RANSAC framework for outliers removal and pose estimation in visual odometry.
The proposed method has been tested on both synthetic data and real-world scenes from KITTI. The experimental results show that the accuracy of the estimated poses is superior to other alternative methods.
\end{abstract}

\begin{IEEEkeywords}
Vision-Based Navigation, Visual-Inertial SLAM, Sensor Fusion.
\end{IEEEkeywords}

\section{INTRODUCTION}
\IEEEPARstart{W}{ith} the rapid development of autonomous navigation, autonomous driving, and augmented reality, the combination of camera and IMU is becoming increasingly popular. The camera offers the benefits of non-contact operation, high precision, miniaturization, cost-effectiveness, and rich information. Meanwhile, IMU can deliver precise short-term position, velocity, acceleration, and angle data solely through internal sensors without external assistance. So there is a high degree of complementarity of the advantages between cameras and IMUs. The combination of cameras and IMUs has become an indispensable equipment for moving platforms, so the fusion technology of cameras and IMUs has received more and more attention~\cite{li20134}. In particular, Carlos Campos et al. have applied vision and IMU fusion technology to the ORB-SLAM3~\cite{campos2021orb} algorithm and its accuracy has improved dramatically compared to ORB-SLAM2~\cite{mur2017orb}.

\begin{figure}[t]    
  \vspace{-5pt}
     \centering    \includegraphics[width=0.9\linewidth]{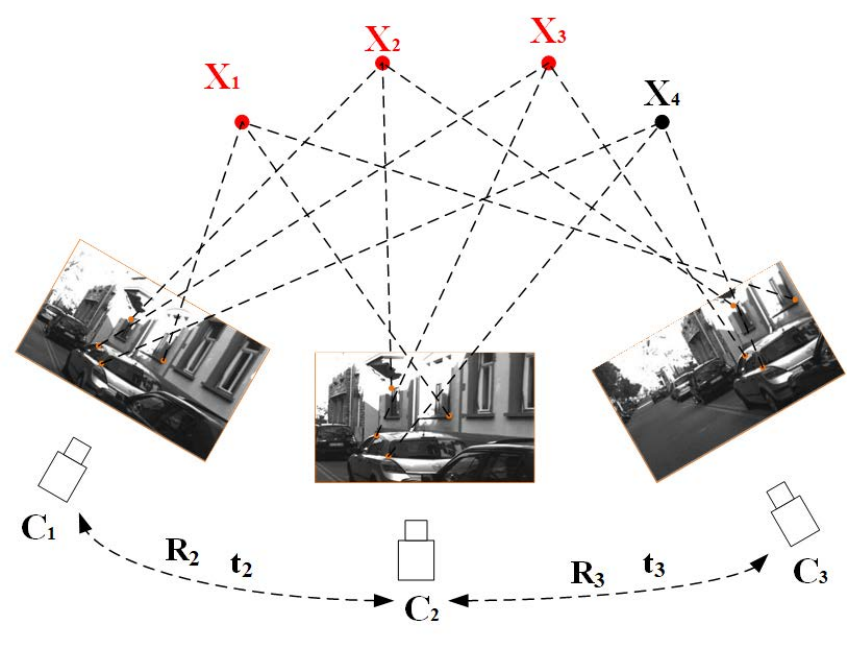}
    \centering
        \vspace{-10pt}
    \caption{Trifocal tensor using 4 point correspondences or 3 point correspondences in three views.}
    \label{fig.0}
       \vspace{-5pt}
\end{figure}

Estimating the relative pose is the most critical step in computer vision fields, including simultaneous localization and mapping (SLAM) and structure from motion (SfM)~\cite{yu2023globally}. The relative pose estimation of two views is already very mature, but there are still many difficulties and challenges in the pose estimation of the 3-view problem, such as the typical 3v4p (3-view 4-point ) perspective pose problem~\cite{nister2006four}, see Fig. \ref{fig.0}. 
But overall, compared with the 2-view pose estimation represented by epipolar constraints, the 3-view pose estimation has the advantage of requiring fewer matching features and undergoing less degeneration~\cite{ding2023minimal}, for its constraints are tighter. Since the solution of the 3-view problem is more complex and time-consuming, it should not be regarded as a replacement or a competitor for 2-view pose estimation, but rather as a fallback option when the 2-view pose estimation fails~\cite{julia2018critical}. 

In practice, there will be mismatched features, so combining the pose estimation algorithms with robust estimation frameworks is necessary. The most commonly used robust framework is the random Sample Consensus (RANSAC)~\cite{fischler1981random}, and its iteration numbers are closely related to the minimum sample size required. In general, the number of iterations increases exponentially as the number of matching features required by the algorithm increases. Therefore, it is of great significance to minimize the number of matches needed by the algorithm, because it can effectively boost the efficiency and robustness of pose estimation. 

The minimum solution plays an important role in addressing geometric vision problems. In the classic 2-view pose estimation problems, the rotation matrix $\mathbf{R}$, and the translation vector $\mathbf{t}$ each consists of 3 unknowns, so 5 unknowns still need to be solved without considering scale ambiguity. The most typical method is Nister's 5-point method~\cite{nister2004efficient}, which applies Gauss-Jordan elimination method to reduce the polynomial equations. 
For the 3-view pose estimation problem, theoretically, 4 point correspondences or 6 line correspondences are enough to solve the camera pose for three views~\cite{hartley2003multiple}.
However, these solvers may produce more spurious solutions due to nonlinear constraints, such as up to 600 solutions with the 6-line method~\cite{holt1994motion}.
Ding proposed to solve the 3-view problem using the HC (homotopy continuation) solver, which needs GPU implementations for parallel calculations, so it is also called GPU-HC solver~\cite{ding2023minimal}.
 Fabbri et al. also proposed two minimum solutions based on the HC   solver~\cite{fabbri2020trplp}.
Inspired by~\cite{guan2022trifocal}, the main contributions of this paper can be summarized as follows.

\begin{itemize}
        \item By fusing IMU information into the 3-view problem, the linear closed-form solution with 4 point correspondences can be derived. Compared with the latest 4-point method, our method can operate in real-time on CPU, which is more efficient, simple, and practical.
        \item We propose a new minimal solver using only 3 point correspondences, which uses fewer point correspondences than other 3-view problem solvers. The minimum problem can be converted into two polynomials with two unknowns about rotation angles among three views. The maximum polynomial degree is 12, and the minimal solver is found with the automatic Gröbner basis solver.
        \item In extensive synthetic and real experiments, we have proved that the proposed methods can achieve higher accuracy than the traditional 3-view problem linear solver, and even higher than the accuracy of the classic 2-view problem solvers.
\end{itemize}

The rest of this paper is arranged as follows. In Section II, the related studies are reviewed. In Section III, the \verb+3pt-Our-3view+ and \verb+4pt-Our-3view+ methods are proposed, and their detailed principles are deduced. In Section IV, the performance of the proposed methods is tested using both synthetic and real-world datasets. Finally, we present the conclusions in Section V.

\section{RELATED WORK}
Since 3-view problem solving is still a hard problem~\cite{nister2006four},
as far as we know, there are few publicly available solvers for the 3-view problem. The problem has 12 degrees of freedom, where the rotation matrix $\mathbf{R}$ and the translation vector $\mathbf{t}$ each consists of 6 unknowns.

\textbf{3-view point-based problems:} 
For the 3-view point-based problem,
Heyden et al. introduced the concept of reduced fundamental tensor and proposed that solving the trifocal relative pose at least needs 6 points~\cite{heyden1997reconstruction}. 
Based on algebraic constraints of the correlation slices, Ressl et al. proposed the minimal parameterization of the trifocal tensor~\cite{ressl2002minimal}.
Nordberg et al. introduced three 3×3 orthogonal matrices which can transform the original trifocal tensors into a sparse form with only 10 non-zero parameters up-to-scale~\cite{nordberg2009minimal}.
Faugeras et al. presented a set of 12 algebraic equations that serve as adequate constraints for characterizing a trifocal tensor~\cite{faugeras1998nonlinear}.
Ponce et al. proposed 6 homogeneous constraints that can be obtained through a projective transformation of the space, and explored a new approach to characterize the 3-view model~\cite{ponce2014trinocular}.
The most relevant paper to our research is~\cite{ding2023minimal}, Ding et al. solved minimal problems with 4 point correspondences in three views and illustrated the effective use of GPU implementations for the Homotopy Continuation solver. The solver needs to train a model to predict a starting solution for the Homotopy Continuation method in order to obtain a satisfactory solution~\cite{hruby2022learning}.
Xu et al. proposed an accurate and real-time point-line based pose estimator that fully exploits the correlation constraints of points and lines, making it able to handle both general and degenerate cases~\cite{xu2024accurate}.

\textbf{Polynomials solvers:} It is well known that pose estimation problems can be converted into polynomial solving.
Kukelovaet al. first presented the automatic Gröbner basis solver, which has been widely used for minimal problems~\cite{kukelova2008automatic}.
However, the Gröbner solver is unstable and difficult to solve large-scale problems, but now with the deepening of related research, the stability and speed of Gröbner basis methods have been further improved~\cite{larsson2018beyond}.
Based on~\cite{kukelova2008automatic}, Larsson et al. proposed a more efficient solver that can exploit intrinsic relations of the input polynomials~\cite{larsson2017efficient}.
Li et al. proposed the GAPS solver, and the efficiency, usability, and flexibility of the solver have all been improved~\cite{li2020gaps}.
Bhayani et al. studied an alternative algebraic method that can convert resultant constraints to eigenvalue problems~\cite{bhayani2020sparse}. 
Martyushev et al. proposed the latest Gröbner basis solver for Laurent polynomial equations, which can check whether the equations meet the requirements for building the elimination template~\cite{martyushev2023automatic}. 
Ding et al. proposed some new insights into minimal solvers based on elimination theory, which helps generate more efficient solvers~\cite{kukelova2017clever}~\cite{ding2021general}.
In addition, the HC solver is also an effective numerical solver for polynomials solving~\cite{sommese2005numerical}. Recent studies have shown that trifocal tensor can be effectively resolved by GPU-HC solver, for it can accelerate the speed in orders of magnitude, but they are still limited by the computer hardware level~\cite{chien2022gpu}. Since the HC solver needs a starting solution and higher hardware configuration, we still use the latest Gröbner basis solver in this paper.

\section{POSE ESTIMATION}

\subsection{Trifocal Tensors}

Based on the camera projection model described in~\cite{hartley2003multiple}, the 
camera's projection matrix can be represented as $\mathbf{P}=\mathbf{K}[\mathbf{R}\mid -\mathbf{R}\tilde{\mathbf{C}}]$, where $\tilde{\mathbf{C}}$ 
is the world coordinate of the camera's center.
Assuming $\mathbf{X}$ is a spatial 3D point, its image coordinate $\mathbf{x}$ can be represented as $\mathbf{x}=\mathbf{P}\mathbf{X}$.
When the camera calibration matrix $\mathbf{K}$ is known, 
 we can obtain its normalized image coordinate $\widehat{\mathbf{x}}=\mathbf{K}^{-1}\mathbf{x}$.
 This implies that $\widehat{\mathbf{x}}$ is the corresponding image point of $\mathbf{X}$ 
 under the camera with the calibration matrix $\mathbf{K}=\mathbf{I}$, where $\mathbf{I}$ denotes the identity matrix.
 In order to remove the influence of camera calibration matrices, we define $\widehat{\mathbf{P}}=\mathbf{K}^{-1}\mathbf{P}=[\mathbf{R}\mid -\mathbf{R}\tilde{\mathbf{C}}]$ 
 as the normalized projection matrix.
 In the following paragraphs, all the camera calibration matrices are assumed to be intrinsically calibrated, and all projection matrices 
 and image coordinates are normalized projection matrices and normalized image coordinates.
 
In the three views, the projection matrix for each camera can be represented as:
\begin{equation}
	\begin{aligned}                \mathbf{P}_k=\begin{bmatrix}1&0&0&0\\0&1&0&0\\0&0&1&0\end{bmatrix}\begin{bmatrix}\mathbf{R}_k&-\mathbf{R}_k\mathbf{t}_k\\\mathbf{0}&1\end{bmatrix},
	\end{aligned}
	\label{a1}
\end{equation}
where $k=1,2,3$, $\mathbf{P}_k$ is the projection matrix of the k-th camera. Assuming that the first view is at the origin of the world coordinate system, i.e. 
$\mathbf{R}_1=\mathbf{I}_{3\times3}$, $\mathbf{t}_1=[0,0,0]^T$,
 $\mathbf{R}_k$ and $\mathbf{t}_k$ denote the rotation matrix and translation vector between the k-th view and the first view, respectively.
 
In order to facilitate the derivation of trifocal tensors, the projection matrices can be expressed as 
 $\mathbf{P}_1=[\mathbf{I}|\mathbf{0}]$, $\mathbf{P}_2=[\mathbf{A}|\mathbf{a}_4]$ and $\mathbf{P}_3=[\mathbf{B}|\mathbf{b}_4]$, respectively.
$\mathbf{A}$ and $\mathbf{B}$ are $3\times3$ matrices, and $\mathbf{a_i}$ and $\mathbf{b_i}$ represent i-th columns of corresponding projection matrices. It is worth noting that $\mathbf{a_4}$ and $\mathbf{b_4}$ are the epipoles generated by the first camera in the second and third views respectively, i.e. $\mathbf{e}^{\prime}$ and $\mathbf{e}^{\prime\prime}$.
Then, the trifocal tensors of the three views can be expressed as:
\begin{equation}
	\begin{aligned} 
\mathbf{T}_i=\mathbf{a}_i\mathbf{b}_4^\mathrm{T}-\mathbf{a}_4\mathbf{b}_i^\mathrm{T},\quad \quad i=1,2,3. 
	\end{aligned}
	\label{a2}
\end{equation}

\begin{figure}[htbp]  
  \centering
  \vspace{-10pt}
    \includegraphics[width=0.9\linewidth]{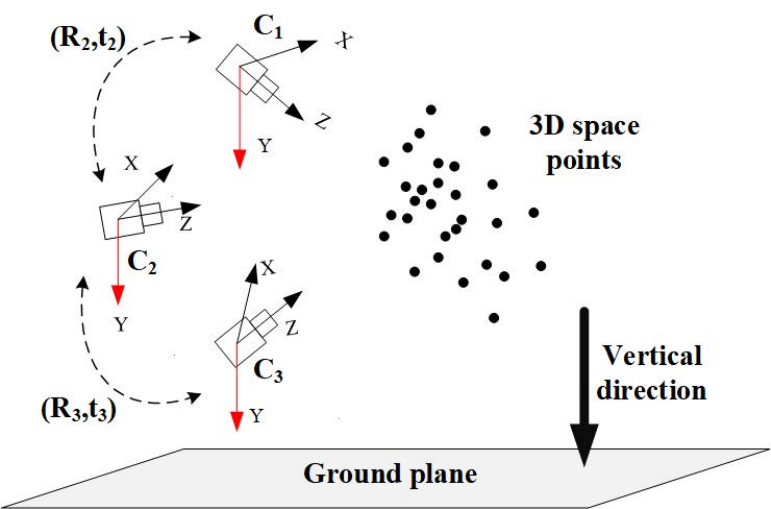}
    \centering
        \vspace{-5pt}
    \caption{Camera alignment with known vertical direction.}
    \label{fig.alien}
        \vspace{-5pt}
\end{figure}

In the common case, the rotation transformation between views belongs to the Euclidean transformation, which can be characterized by yaw, pitch, and roll angles. Using the measurement information of IMUs, the pitch and roll angles of each view can be obtained. Then, the camera's y-axis can be aligned to the direction of gravity, i.e., the y-axis is perpendicular to the ground plane, see Fig. \ref{fig.alien}. 

Assuming $\mathbf{\tilde{x}}^k$ represents the image point, $\mathbf{R}_{\text{imu}}$ denotes the alignment matrix of k-th view. After alignment, its aligned image point $\mathbf{x}^k$ can be represented as:
\begin{equation}
\begin{aligned}
\mathbf{x}^k=\mathbf{R}_{\text{imu}}\mathbf{\tilde{x}}^k.
\end{aligned}
\label{b2}
\end{equation}
Similarly, in the following paragraphs, all the image points are pre-aligned, i.e., aligned image points. Then, the rotation transformation $\mathbf{R}_k$ is only related to the yaw angle, the pose transformation after alignment can be expressed as:
\begin{equation}
	\begin{aligned} 
\mathbf{R}_k=\begin{bmatrix}C_y^k&0&S_y^k\\0&1&0\\-S_y^k&0&C_y^k\end{bmatrix},\mathbf{t}_k=\begin{bmatrix}t_x^k\\t_y^k\\t_z^k\end{bmatrix},
	\end{aligned}
	\label{b3}
\end{equation}
where $\theta_k$ represents the yaw angle of k-th view and $C_y^k=\cos(\theta_k)$, $S_y^k=\sin(\theta_k)$.  

Conversely, the original relative pose transformation $\mathbf{\tilde{R}}_k$ and $\mathbf{\tilde{t}}_k$ can
be recovered from the following equations:
\begin{equation}
	\begin{aligned} 
\begin{cases}\tilde{\mathbf{R}}_k=(\mathbf{R}_{\mathrm{imu}})^T\mathbf{R}_k\mathbf{R}_{\mathrm{imu}}^{\prime}\\\tilde{\mathbf{t}}_k=(\mathbf{R}_{\mathrm{imu}})^T\mathbf{t}_k\end{cases},
	\end{aligned}
	\label{b4}
\end{equation}
where $\mathbf{R}_{\mathrm{imu}}^{\prime}$ represents the alignment matrix of the first view, $\mathbf{R}_{\text{imu}}$ represents the alignment matrix of k-th view. 

\subsection{The 4-point Case} 
Substituting \eqref{b3} into \eqref{a1}, the projection matrix can be expressed as:
\begin{equation}
	\begin{aligned} 
\mathbf{P}_k=\begin{bmatrix}C_y^k&0&S_y^k&-C_y^kt_x^k-S_y^kt_z^k\\0&1&0&-t_y^i\\-S_y^k&0&C_y^k&S_y^kt_x^k-C_y^kt_z^k\end{bmatrix}.
	\end{aligned}
	\label{c1}
\end{equation}

Then, substituting $\mathbf{a_i}$ and $\mathbf{b_i}$ from columns of $\mathbf{P}_k$ into \eqref{a2}, we can deduce the trifocal tensors as follows:
\begin{equation}
	\begin{aligned} 
\mathbf{T}_1=\begin{bmatrix}Q_1&Q_2&Q_3\\Q_4&0&Q_5\\Q_6&Q_7&Q_8\end{bmatrix},
	\end{aligned}
	\label{c2}
\end{equation}
\begin{equation}
	\begin{aligned} 
\mathbf{T}_2=\begin{bmatrix}0&Q_9&0\\Q_{10}&Q_{11}&Q_{12}\\0&Q_{13}&0\end{bmatrix},
	\end{aligned}
	\label{c3}
\end{equation}
\begin{equation}
	\begin{aligned} 
\mathbf{T}_3=\begin{bmatrix}Q_{14}&-Q_7&Q_{15}\\-Q_5&0&Q_4\\Q_{16}&Q_2&Q_{17}\end{bmatrix},
	\end{aligned}
	\label{c4}
\end{equation}
where
\begin{equation}
	\begin{aligned} 
\begin{cases}Q_1=C_y^3Q_9+C_y^2Q_{10,}&Q_2=-C_y^2t_y^3,\\Q_3=C_y^2Q_{12}-S_y^3Q_9,&Q_4=C_y^3t_y^2,\\Q_5=-S_y^3t_y^2,&Q_6=-S_y^2Q_{10}+C_y^3Q_{13},\\Q_7=S_y^2t_y^3,&Q_8=-S_y^2Q_{12}-S_y^3Q_{13},\\Q_9=C_y^2t_x^2+S_y^2t_z^2,&Q_{10}=-C_y^3t_x^3-S_y^3t_z^3,\\Q_{11}=t_y^2-t_y^3,&Q_{12}=S_y^3t_x^3-C_y^3t_z^3,\\Q_{13}=C_y^2t_z^2-S_y^2t_x^2,&Q_{14}=S_y^3Q_9+S_y^2Q_{10},\\Q_{15}=S_y^2Q_{12}+C_y^3Q_9,&Q_{16}=S_y^3Q_{13}+C_y^2Q_{10},\\Q_{17}=C_y^2Q_{12}+C_y^3Q_{13},\end{cases}.
	\end{aligned}
	\label{c5}
\end{equation}

Considering a set of point correspondence $\mathbf{x}^1\leftrightarrow\mathbf{x}^2\leftrightarrow\mathbf{x}^3$ from three views, its constraints can be expressed as:
\begin{equation}
	\begin{aligned} 
\left[\mathbf{x}^2\right]_\times\left(\sum_ix^1_i\mathbf{T}_i\right)\left[\mathbf{x}^3\right]_\times=\mathbf{0}_{3\times3}, 
	\end{aligned}
	\label{c6}
\end{equation}
where $\mathbf{x}^1=[x^1_x,x^1_y,x^1_z]$, $x^1_i$ denotes the i-th coordinate of $\mathbf{x}^1$. The corresponding image coordinates of view 2 and view 3 are denoted as $\mathbf{x}^2$ and $\mathbf{x}^3$, and $\mathbf{T}_i$ denotes the i-th matrix of the trifocal tensor. 

Although Eq. \eqref{c6} contains 9 equations, after analysis, we find that only 4 equations are independent. 
When there are 4 point correspondences, 16 independent equations can be constructed.
Separating $[Q_1\quad Q_2\quad...\quad Q_{17}]^T$ from the equations, the system of 16 equations can be expanded into matrix representations,
\begin{equation}
	\begin{aligned} 
\begin{bmatrix}f_1^\mathrm{l}&f_2^\mathrm{l}&\cdots&\cdots&f_{17}^\mathrm{l}\\f_1^2&f_1^2&\cdots&\dots&f_{17}^2\\&&\cdots&\cdots&&\\f_1^{16}&f_2^{16}&\cdots&\cdots&f_{17}^{16}\end{bmatrix}\begin{bmatrix}Q_\mathrm{l}\\Q_\mathrm{l}\\\cdots\\\cdots\\Q_{17}\end{bmatrix}=0_{16\times1}.
	\end{aligned}
	\label{c7}
\end{equation}

The Eq. \eqref{c7} is in the form of $\mathbf{A}\mathbf{q}=0$, and $\mathbf{q}$ is up to scale. 
Every element $f_m^n$ in $\mathbf{A}$ is known, where $n$ represents the n-th independent equation, 
and $m$ represents the coefficient of $Q_m$.
By setting $\|\mathbf{q}\|=1$, 
and using SVD decomposition, we can solve $[Q_1\quad Q_2\quad...\quad Q_{17}]^T$. Then according to Eq. \eqref{c5}, $\theta_2$, $\mathbf{t}_2$, $\theta_3$, $\mathbf{t}_3$ can be calculated. The results can be further optimized using fixed constraints, and more details can be found in section III-D.

\begin{itemize}
        \item Calculate $\theta_2$:
        \begin{equation}
	\begin{aligned} 
        \begin{cases}S_y^2=(Q_8Q_9+Q_{13}Q_{14})/(Q_{10}Q_{13}-Q_9Q_{12})\\C_y^2=(Q_1Q_{13}-Q_9Q_{17})/(Q_{10}Q_{13}-Q_9Q_{12})\\\theta_2=\arctan2(S_y^2,C_y^2)\end{cases}.
        \end{aligned}
	\label{c8}
        \end{equation}
        \item Calculate $\mathbf{t}_2$:
        \begin{equation}
	\begin{aligned}
        \mathbf{t}_2=\begin{bmatrix}C_y^2Q_9-S_y^2Q_{13}\\C_y^3Q_4-S_y^3Q_5\\S_y^2Q_9+C_y^2Q_{13}\end{bmatrix}.
        \end{aligned}
	\label{c9}
        \end{equation}
        \item Calculate $\theta_3$:
        \begin{equation}
	\begin{aligned} 
        \begin{cases}S_y^3=-(Q_8+S_y^2Q_{12})/Q_{13}\\C_y^3=(Q_1-C_y^2Q_{10})/Q_9\\\theta_3=\arctan2(S_y^3,C_y^3)\end{cases}. 
        \end{aligned}
	\label{c10}
        \end{equation}
        \item Calculate $\mathbf{t}_3$:
        \begin{equation}
	\begin{aligned} 
        \mathbf{t}_3=\begin{bmatrix}S_y^3Q_{12}-C_y^3Q_{10}\\S_y^2Q_7-C_y^2Q_2\\-S_y^3Q_{10}-C_y^3Q_{12}\end{bmatrix}.
        \end{aligned}
	\label{c11}
        \end{equation}
\end{itemize}

After solving for $\theta_2$, $\mathbf{t}_2$, $\theta_3$, $\mathbf{t}_3$, the relative pose estimation before vertical direction correction among the three views can
be further recovered by using Eq. \eqref{b4}.
\subsection{The 3-point Case}
Unlike Eq. \eqref{b3}, $\mathbf{R}_k$  here takes the form of Cayley parameterization, and the form of $\mathbf{t}_k$ remains unchanged.
\begin{equation}
	\begin{aligned} 
\mathbf{R}_k=\frac1{1+{s_k}^2}\begin{bmatrix}1-{s_k}^2&0&2s_k\\0&1+{s_k}^2&0\\-2s_k&0&1-{s_k}^2\end{bmatrix}, \mathbf{t}_k=\begin{bmatrix}t_x^k\\t_y^k\\t_z^k\end{bmatrix}, 
\end{aligned}
\label{d1}
\end{equation}
where $\theta_k$ represents the yaw angle, $s_k=\tan(\theta_k/2)$. In practice, the yaw angle is rarely equal to 180°, so this situation is almost negligible~\cite{larsson2017efficient}.

Substituting Eq. \eqref{d1} into $\mathbf{P}_k$, and combining 
 Eq. \eqref{a2}, then every equation in \eqref{c6} can be expressed as :
\begin{equation}
	\begin{aligned} 
\begin{aligned}
        &\begin{aligned}\frac{f_1^i(s_2,s_3)t_{2x}+f_2^i(s_2,s_3)t_{2y}+f_3^i(s_2,s_3)t_{2z}}{(1+s_2^2)(1+s_3^2)}\end{aligned} \\
        &\begin{aligned}+\frac{f_4^i(s_2,s_3)t_{3x}+f_5^i(s_2,s_3)t_{3y}+f_6^i(s_2,s_3)t_{3z}}{(1+s_2^2)(1+s_3^2)}&=0\end{aligned}
        \end{aligned}, 
\end{aligned}
\label{d3}
\end{equation}
where $f_{\star}^{i}(s_2,s_3)$ is a polynomial with unknowns limited to $s_2$ and $s_3$. 
$i$ represents the i-th equation, $i=1,..,9$. 
Separating $[t_{2x}\quad t_{2y}\quad t_{2z}\quad t_{3x}\quad t_{3y}\quad t_{3z}]^T$ from \eqref{d3}, and define it as $\mathbf{t}$, the equations can be expanded into matrix representations.
\begin{equation}
\begin{aligned} 
&\frac1{(1+s_2^2)(1+s_3^2)}\times\mathbf{F}(s_2,s_3) \mathbf{t}=\mathbf{0}_{6\times1},
\end{aligned}
\label{d4}
\end{equation}
where 
\begin{equation}
\begin{aligned} 
\mathbf{F}(s_2,s_3)=\begin{bmatrix}f_1^1&f_2^1&f_3^1&f_4^1&f_5^1&f_6^1\\f_1^2&f_2^2&f_3^2&f_4^2&f_5^2&f_6^2\\&&&\cdots&\cdots&&\\f_1^9&f_2^9&f_3^9&f_4^9&f_5^9&f_6^9\end{bmatrix}.
\end{aligned}
\label{d41}
\end{equation}
Although the dimensionality of $\mathbf{F}(s_{2},s_{3})$ is ${9\times6}$, only 3 rows are independent. When there are 3 point correspondences, 27 equations can be constructed. 
By selecting independent rows of each point match, 9 equations can be constructed, and its matrix representations can be expressed as:
\begin{equation}
	\begin{aligned} 
\mathbf{\widehat{F}}\mathbf{t}=\mathbf{0}_{6\times1}, 
\end{aligned}
\label{d5}
\end{equation}
where 
\begin{equation}
\begin{aligned} 
\mathbf{\widehat{F}}=\begin{bmatrix}f_1^1&f_2^1&f_3^1&f_4^1&f_5^1&f_6^1\\f_1^2&f_2^2&f_3^2&f_4^2&f_5^2&f_6^2\\f_1^4&f_2^4&f_3^4&f_4^4&f_5^4&f_6^4\\f_1^{10}&f_2^{10}&f_3^{10}&f_4^{10}&f_5^{10}&f_6^{10}\\f_1^{11}&f_2^{11}&f_3^{11}&f_4^{11}&f_5^{11}&f_6^{11}\\f_1^{13}&f_2^{13}&f_3^{13}&f_4^{13}&f_5^{13}&f_6^{13}\\f_1^{19}&f_2^{19}&f_3^{19}&f_4^{19}&f_5^{19}&f_6^{19}\\f_1^{20}&f_2^{20}&f_3^{20}&f_4^{20}&f_5^{20}&f_6^{20}\\f_1^{22}&f_2^{22}&f_3^{22}&f_4^{22}&f_5^{22}&f_6^{22}\end{bmatrix}
\end{aligned}.
\label{d51}
\end{equation}
The superscript of $f_{\star}^{i}$ is the i-th equation from 27 equations. According to the row sequences, every 3 rows are independent and correspond to a point match, such as rows 1, 2, and 4 corresponding to the first point match. Note that the multiple $1/(1+s_2^2)(1+s_3^2)$ has been omitted in \eqref{d5}, but the 
multiple is still useful in the later paper. It can be used to reduce the degree of the polynomial in the determinant and enhance the solution efficiency.

Since there is a feasible solution $\mathbf{t}$, any 6 rows taken from $\mathbf{\widehat{F}}$ can form a new matrix, and its determinant is 0,
\begin{equation}
	\begin{aligned} 
F_1(s_2,s_3)=\begin{vmatrix}f_1^1&f_2^1&f_3^1&f_4^1&f_5^1&f_6^1\\f_1^2&f_2^2&f_3^2&f_4^2&f_5^2&f_6^2\\f_1^4&f_2^4&f_3^4&f_4^4&f_5^4&f_6^4\\f_1^{10}&f_2^{10}&f_3^{10}&f_4^{10}&f_5^{10}&f_6^{10}\\f_1^{11}&f_2^{11}&f_3^{11}&f_4^{11}&f_5^{11}&f_6^{11}\\f_1^{13}&f_2^{13}&f_3^{13}&f_4^{13}&f_5^{13}&f_6^{13}\end{vmatrix}_{6\times6}=0.
\end{aligned}
\label{d6}
\end{equation}
Then, we can obtain multiple equations similar to \eqref{d6}, which have 2 unknowns including $s_2$, $s_3$. In the normal case, the equations can be solved, but it is not easy. Because \eqref{d6} is of degree 24, solving high-degree equations is still a hard problem. Inspired by ~\cite{guan2023minimal}, we use the multiple $1/(1+s_2^2)(1+s_3^2)$ to reduce the degree of the polynomial, the equations can be simplified as follows:
\begin{equation}
	\begin{aligned} 
\mathrm{quot}\left(\sum_{i=0}^{12}\sum_{j=0}^{12}w_{ij}s_2^is_3^j,(1+s_2^2)^3(1+s_3^2)^3\right)=0,
\end{aligned}
\label{d8}
\end{equation}
where $\mathrm{quot}(a,b)$ denotes the quotient of $a$ divided
by $b$. By simplifying, the degree of $s_2$ and $s_3$ is reduced to 6, and the total degree of the equations can be reduced to half of their original degree, i.e. the equations are of degree 12.
\begin{itemize}

        \item Calculate $s_2$ and $s_3$:
        
        Using the latest Gröbner Automatic Solver proposed by~\cite{li2020gaps}, we can solve \eqref{d8} accurately. Even though Eq. \eqref{d8} has been simplified, the equation degree is still very high. As a result, the solution process is time-consuming. The Solver would obtain 6 complex roots, from which we choose all the real roots as possible solutions for $s_2$ and $s_3$.
        \item Calculate $\mathbf{t}$:

        After $s_2$ and $s_3$ are solved, we substitute $s_2$ and $s_3$ into $\mathbf{\widehat{F}}$ in \eqref{d5}. By setting $|\mathbf{t}|=1$, 
        and using SVD decomposition, we can solve $[t_{2x}\quad t_{2y}\quad  t_{2z}\quad  t_{3x}\quad  t_{3y}\quad  t_{3z}]^T$.

\end{itemize}

\subsection{Constraint Enforcement}
No matter the 4-point case or the 3-point case, we can both get the trifocal tensor $\mathbf{T}_i$, but it does not strictly satisfy the constraints of Eq. \eqref{a2}, so it is necessary to enforce constraints on $\mathbf{T}_i$ to improve its accuracy further.

Assuming $u_i$ and $v_i$ are the left and right null-vectors of $\mathbf{T}_i$ respectively, i.e. $\mathbf{u}_i^\mathsf{T}\mathbf{T}_i=\mathbf{0}^\mathsf{T}$, $\mathrm{T}_{i}\mathbf{v}_{i}=\mathbf{0}$, the epipoles $\mathbf{e}^{\prime}$ and $\mathbf{e}^{\prime\prime}$ are perpendicular to the three left and right null-vectors of $\mathbf{T}_i$. Using SVD decomposition, the epipoles can be obtained.
\begin{equation}
\begin{cases}\mathrm{e}^{\prime\mathsf{T}}[\mathbf{u}_1,\mathbf{u}_2,\mathbf{u}_3]=\mathbf{0}\\\mathbf{e}^{\prime\prime\mathsf{T}}[\mathbf{v}_1,\mathbf{v}_2,\mathbf{v}_3]=\mathbf{0}\end{cases}.
	\label{f1}
\end{equation}

According to Eq. \eqref{c1} and the obtained epipoles, the projection matrices $P_2$ and $P_3$ can be expressed as follows:
\begin{equation} 
\mathbf{P}_2=\begin{bmatrix}a_1&0&a_2&e_1^{\prime}\\0&a_3&0&e_2^{\prime}\\-a_2&0&a_1&e_3^{\prime}\end{bmatrix},
	\label{f2}
\end{equation}
\begin{equation}
\mathbf{P}_3=\begin{bmatrix}a_4&0&a_5&e_1^{\prime\prime}\\0&a_6&0&e_2^{\prime\prime}\\-a_5&0&a_4&e_3^{\prime\prime}\end{bmatrix},
	\label{f3}
\end{equation}
where $a_*$ represents the unknowns, $e^{\prime}_i$ or $e^{\prime\prime}_i$ is the i-th element of epipoles. Then using Eq. \eqref{a2}, we can get the trifocal tensor $\mathbf{T}_i$.
\begin{equation}
\mathbf{T}_1=\begin{bmatrix}-a_4e_1'+a_1e_1''&a_1e_2''&a_5e_1'+a_1e_3''\\-a_4e_2'&0&a_5e_2'\\-a_4e_3'-a_2e_1''&-a_2e_2''&a_5e_3'-a_2e_3''\end{bmatrix},
	\label{f4}
\end{equation}
\begin{equation}
\mathbf{T}_2=\begin{bmatrix}0&-a_6e_1'&0\\a_3e_1''&-a_6e_2'+a_3e_2''&a_3e_3''\\0&-a_6e_3'&0\end{bmatrix},
	\label{f5}
\end{equation}

\begin{equation}
\mathbf{T}_3=\begin{bmatrix}-a_5e_1'+a_2e_1''&a_2e_2''&-a_4e_1'+a_2e_3''\\-a_5e_2'&0&-a_4e_2'\\-a_5e_3'+a_1e_1''&a_1e_2''&-a_4e_3'+a_1e_3''\end{bmatrix}.
	\label{f6}
\end{equation}

Comparing Eq. \eqref{f4} to \eqref{f6} with Eq. \eqref{c2} to \eqref{c4}, we can find that $Q_*$ in Eq. \eqref{c2} to Eq. \eqref{c4} can be expressed as expressions of unknowns $a_*$, its linear relationships can be expressed as:
\begin{equation}
\mathbf{q}=\mathbf{E}\mathbf{a}.
	\label{f61}
\end{equation}
The main advantage of Eq. \eqref{f4} to \eqref{f6} is that they naturally satisfy the constraint in Eq. \eqref{a2}. Then the problem in Eq. \eqref{c7}, i.e. $\mathbf{A}\mathbf{q}=0$ can be transformed into the following form:
\begin{equation}
\mathbf{a}=\arg\min_{\mathbf{a}}\left(\left\|\mathbf{AEa}\right\|\right)\\s.t.\left\|\mathbf{Ea}\right\|=1
	\label{f7}
\end{equation}
After $\mathbf{a}$ is solved, then using Eq. \eqref{f4} to \eqref{f6} solve the trifocal tensor $\mathbf{T}_i$, and its results are more accurate. 
All the above content outlines our entire process, other methods~\cite{kukelova2017clever}~\cite{ding2021general} may also be viable and require further exploration.

\section{EXPERIMENTS}

In order to validate the accuracy and feasibility of our method, both synthetic and real-world experiments are independently conducted in the paper. Since most 3-view problem solvers do not have open source code, or some have code but no comparison, we have not found an effective way to compare our method with the latest solver~\cite{ding2023minimal} yet. So, except for the public 3-view normalized linear solution~\cite{hartley2003multiple}, we also compare our method with some classic mature 2-view algorithms, including \verb+8pt-Hartley-2view+~\cite{hartley2003multiple}, and \verb+5pt-Nister-2view+~\cite{nister2004efficient} methods, which can also be used for the 3-view pose estimation. 
Different from the method~\cite{ding2023minimal}, which requires GPU implementations, the above methods are all based on CPU implementations.
In this paper, our proposed methods are referred to as \verb+3pt-Our-3view+ and \verb+4pt-Our-3view+, the ordinary normalized solution for the 3-view problem is referred to as \verb+7pt-Hartley-3view+, and its code can be found in~\cite{julia2018critical}. Since 2-view and 3-view are different and each has its own advantages, our experiments are not intended to prove that our methods outperform 2-view algorithms, but to illustrate the feasibility and practicality of the proposed methods.

In the experiment, the rotation and translation errors are calculated as follows:
\begin{equation}
	\begin{aligned} 
\varepsilon_{\mathbf{R}}=\arccos((\mathrm{trace}(\mathbf{R}_{gt}\mathbf{R}^T)-1)/2), 
\end{aligned}
\label{e1}
\end{equation}
\begin{equation}
	\begin{aligned} 
\mathbf{\varepsilon_{t}}=\arccos((\mathbf{t}_{gt}^T\mathbf{t})/(\|\mathbf{t}_{gt}\|\cdot\|\mathbf{t}\|)), 
\end{aligned}
\label{e2}
\end{equation}
where $\mathbf{R}_{gt}$ and $\mathbf{t}_{gt}$ are the ground truth values, $\mathbf{R}$ and $\mathbf{t}$ are the estimated values. Since the estimated translation of the monocular camera is up to scale, we adopt the angle metric to measure the translation errors instead of using Euclidean distance. After $\varepsilon_{\mathbf{R}}$ and $\mathbf{\varepsilon_{t}}$ among three views are solved, we choose the median values as the final results.

\subsection{Efficiency Comparison and Numerical Stability}

All methods run on the same computer, i.e. Intel(R) Xeon(R) Gold6133 2.50GHz using MATLAB. The detailed run times are shown in Table \ref{tab: runtime}, we can see that the proposed \verb+4pt-Our-3view+ method has a runtime of 0.95 ms,
 which is longer than the 0.63 ms of the \verb+8pt-Hartley-2view+ method, but it can solve two sets of camera poses at the same time, while the 8-point method only solves one set, so it is more efficient than the \verb+8pt-Hartley-2view+ method. The \verb+4pt-Our-3view+ method runs 4 times faster than the \verb+7pt-Hartley-3view+ method and is the most efficient algorithm in the 3-view problem. 
The \verb+3pt-Our-3view+ method adopts the Gröbner-based approach, and its polynomial coefficients are quite complex. However, it can still operate in real-time and runs 1.4 times faster than the \verb+7pt-Hartley-3view+ method.
\begin{table}[htbp]
\setlength{\tabcolsep}{1mm}
    \caption{Run-time comparison of different methods(unit:ms)}
    \setlength\tabcolsep{4pt}
    \label{tab: runtime}
    \begin{center}
        \vspace{-10pt}
\resizebox{0.48\textwidth}{!}{\begin{tabular}{|c|c|c|c|c|c|}
\hline
Methods & \begin{tabular}[c]{@{}c@{}}8pt-Hartley\\ -2view\end{tabular} & \begin{tabular}[c]{@{}c@{}}5pt-Nister\\ -2view\end{tabular} & \begin{tabular}[c]{@{}c@{}}7pt-Hartley\\ -3view\end{tabular} & \begin{tabular}[c]{@{}c@{}}3pt-Our\\ -3view\end{tabular} & \begin{tabular}[c]{@{}c@{}}4pt-Our\\ -3view\end{tabular} \\ \hline
Time   & 0.63                                                         & 1.46                                                     & 4.24                                                         & 2.93                                            & 0.95 \\ 
\hline
\end{tabular}
}
    \end{center}
        \vspace{-10pt}
 \end{table}
 
\begin{figure}[h]      
\vspace{-20pt}  
  \centering
     \subfloat[${\varepsilon _{\bf{R}}}$]{
     \centering
     \includegraphics[width=0.45\linewidth]{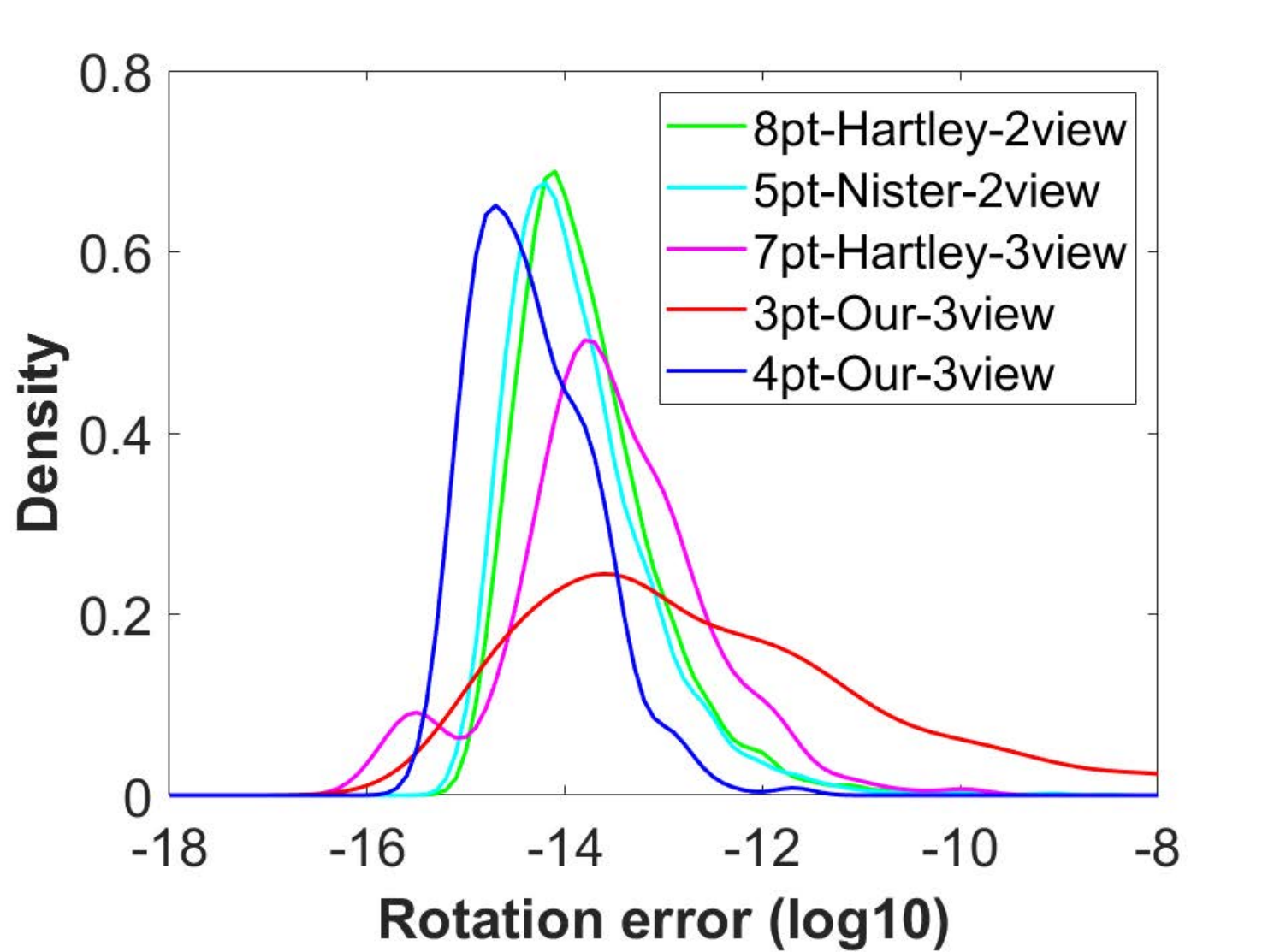}
     \label{fig.1a}
      }
   \subfloat[${\varepsilon _{\bf{t}}}$]{
     \centering
     \includegraphics[width=0.45\linewidth]{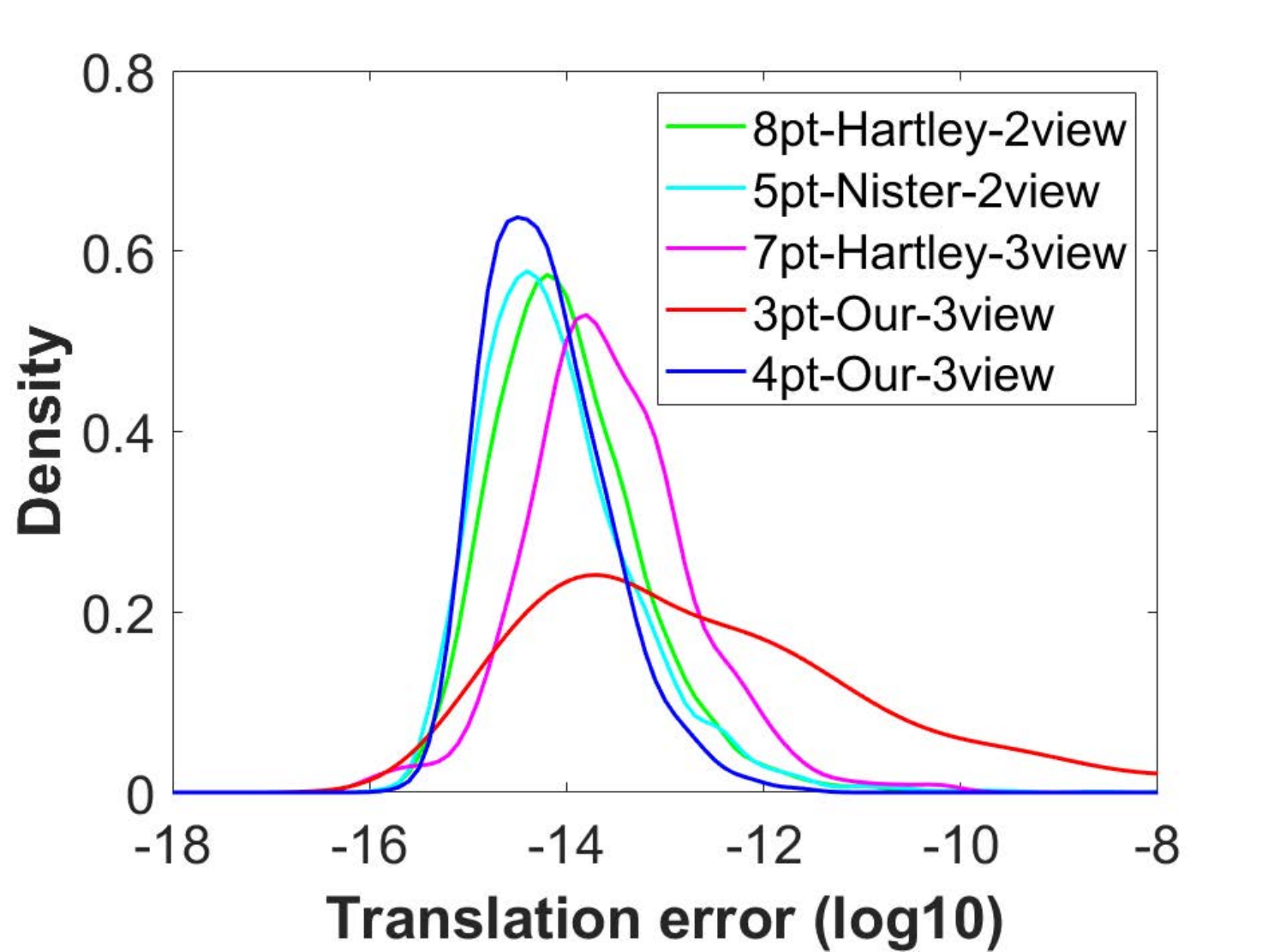}
     \label{fig.1b}
      }
          \centering
    \caption{Probability density functions over relative pose estimation errors. (a) Rotation estimation errors. (b)Translation estimation errors. The horizontal axis represents the $log_{10}$ value of the estimated error, and the vertical axis represents the probability density. The narrower the curve, the smaller the data variance, indicating that the algorithm is more stable}.
    \label{fig.1}
    \vspace{-10pt}
\end{figure}

Numerical stability refers to the ability of an algorithm to provide consistent, accurate, and reliable pose estimation under ideal or perfect conditions. Without considering the influence of image noises, the detailed results regarding numerical stability are shown in Fig. \ref{fig.1}. All the methods are executed 5,000 times. The results show that the \verb+4pt-Our-3view+ method is the most stable, followed by the \verb+5pt-Nister-2view+ and \verb+8pt-Hartley-2view+ methods. Although the proposed \verb+3pt-Our-3view+ method has the worst stability, its mean error is less than 1e-13, and most errors are less than 1e-8, so the \verb+3pt-Our-3view+ method is still accurate and feasible.
\subsection{Experiments on Synthetic Data}

In the synthetic experiments, the image resolution is $640 \times 480$ pixels, and its principal point is $(320, 240)$. The rotation angles among the three views are set within a range of -10° to 10°, and the distances between each other are set within a range of -10 m to 10 m. 3D points in space are randomly generated and can be observed by three views simultaneously. In the simulation, the effects of image noise and IMU angle noise on each method were studied separately. All the methods were executed 1000 times.

\subsubsection{Accuracy with image noise}
\begin{figure}[h]
 \vspace{-20pt}
  \centering
     \subfloat[${\varepsilon _{\bf{R}}}$ - Image noise]{
     \centering
     \includegraphics[width=0.45\linewidth]{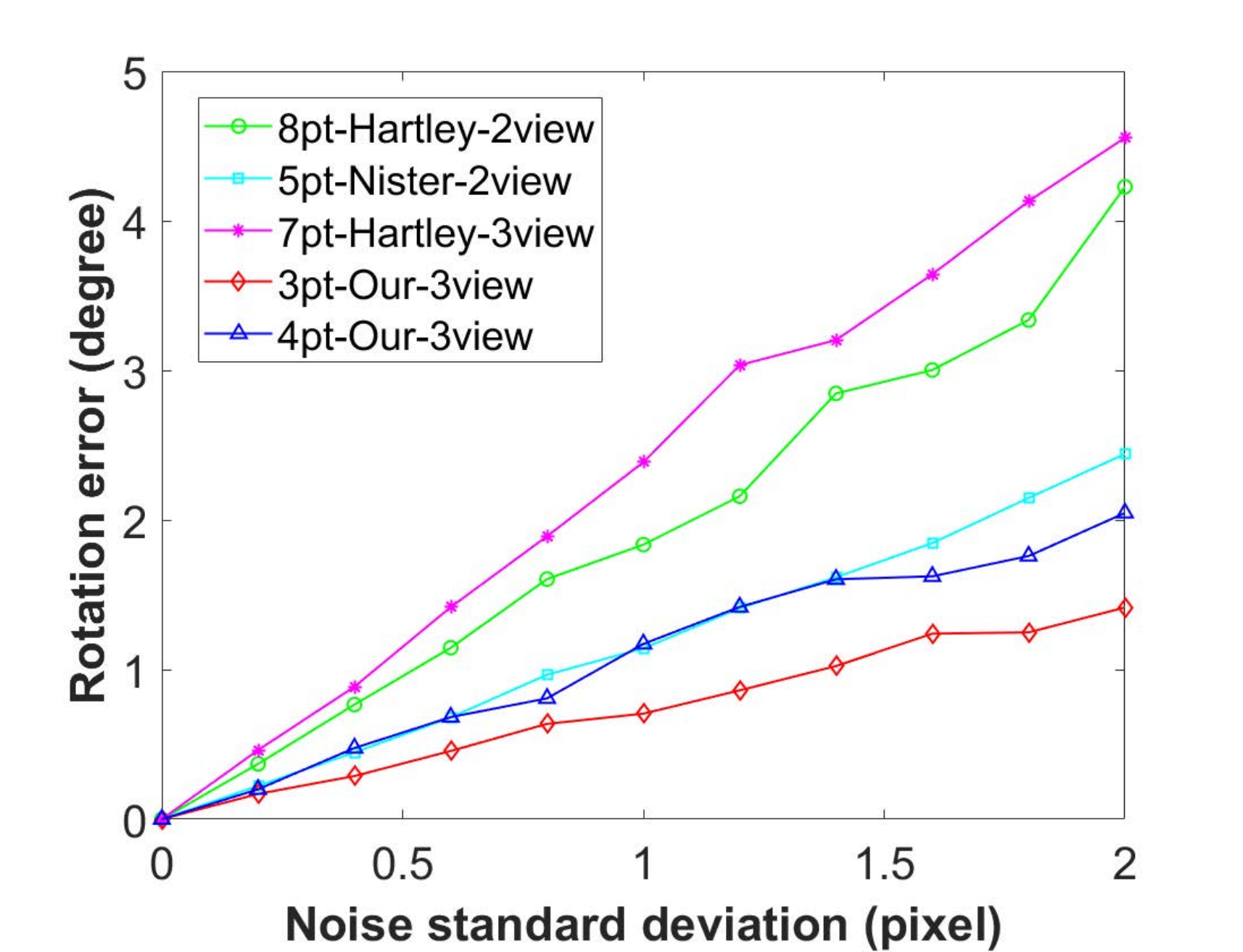}
     \label{fig.image noise a}
      }
   \subfloat[${\varepsilon _{\bf{t}}}$ - Image noise]{
     \centering
     \includegraphics[width=0.45\linewidth]{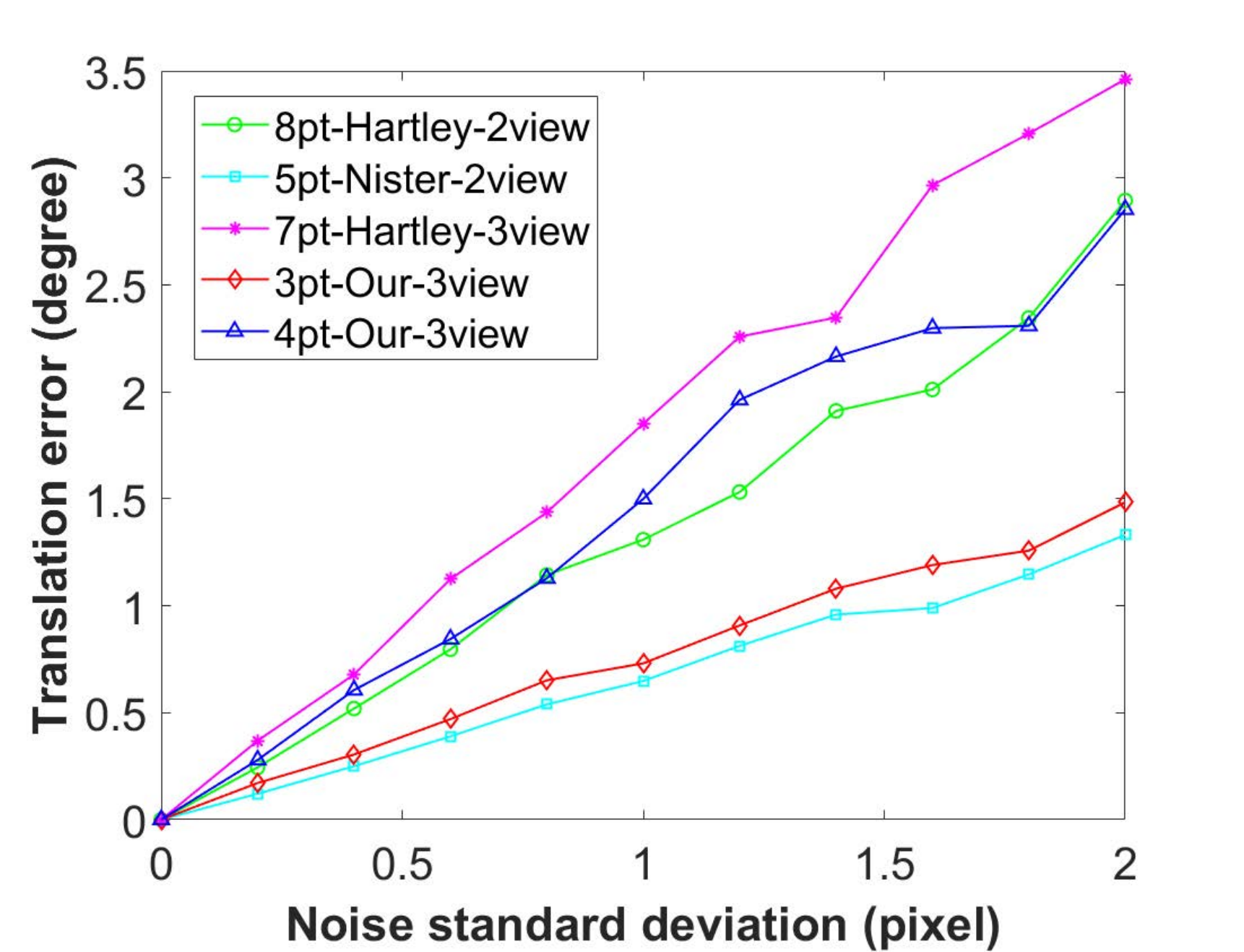}
     \label{fig.image noise b}
      }
          \centering
    \caption{Methods accuracy with image noise. (a) Rotation estimation errors. (b)Translation estimation errors.}
    \label{fig. image noise}
    \vspace{-5pt}
\end{figure}

To study the effects of image noise on the algorithm, in the simulation, we applied Gaussian noise varying from 0 to 2 pixels. The results in Fig. \ref{fig. image noise} show that the \verb+3pt-Our-3view+ method performs best when solving rotation matrices, followed by \verb+4pt-Our-3view+ and \verb+5pt-Nister-2view+ methods. As for solving translation vectors, the \verb+5pt-Nister-2view+ method demonstrates the highest level of accuracy. Despite this, the accuracy of the \verb+3pt-Our-3view+ method also performs well with accuracy almost close to the \verb+5pt-Nister-2view+ method, and has the advantage of solving more parameters with fewer point correspondences.
The \verb+7pt-Hartley-3view+ method performs worst and is susceptible to image noise.
As for the 3-view problem, compared with \verb+7pt-Hartley-3view+, the advantages of \verb+3pt-Our-3view+ and \verb+4pt-Our-3view+ are obvious.
\subsubsection{Accuracy with IMU pitch angle noise}
\begin{figure}[hbp]          
  \centering
     \centering
\includegraphics[width=0.7\linewidth]{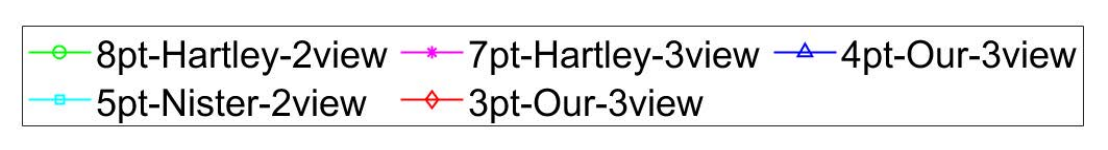}
     \label{fig.IMU angle noise a}
            \vspace{-10pt}
     \subfloat[${\varepsilon _{\bf{R}}}$ - IMU pitch]{
     \centering
     \includegraphics[width=0.45\linewidth]{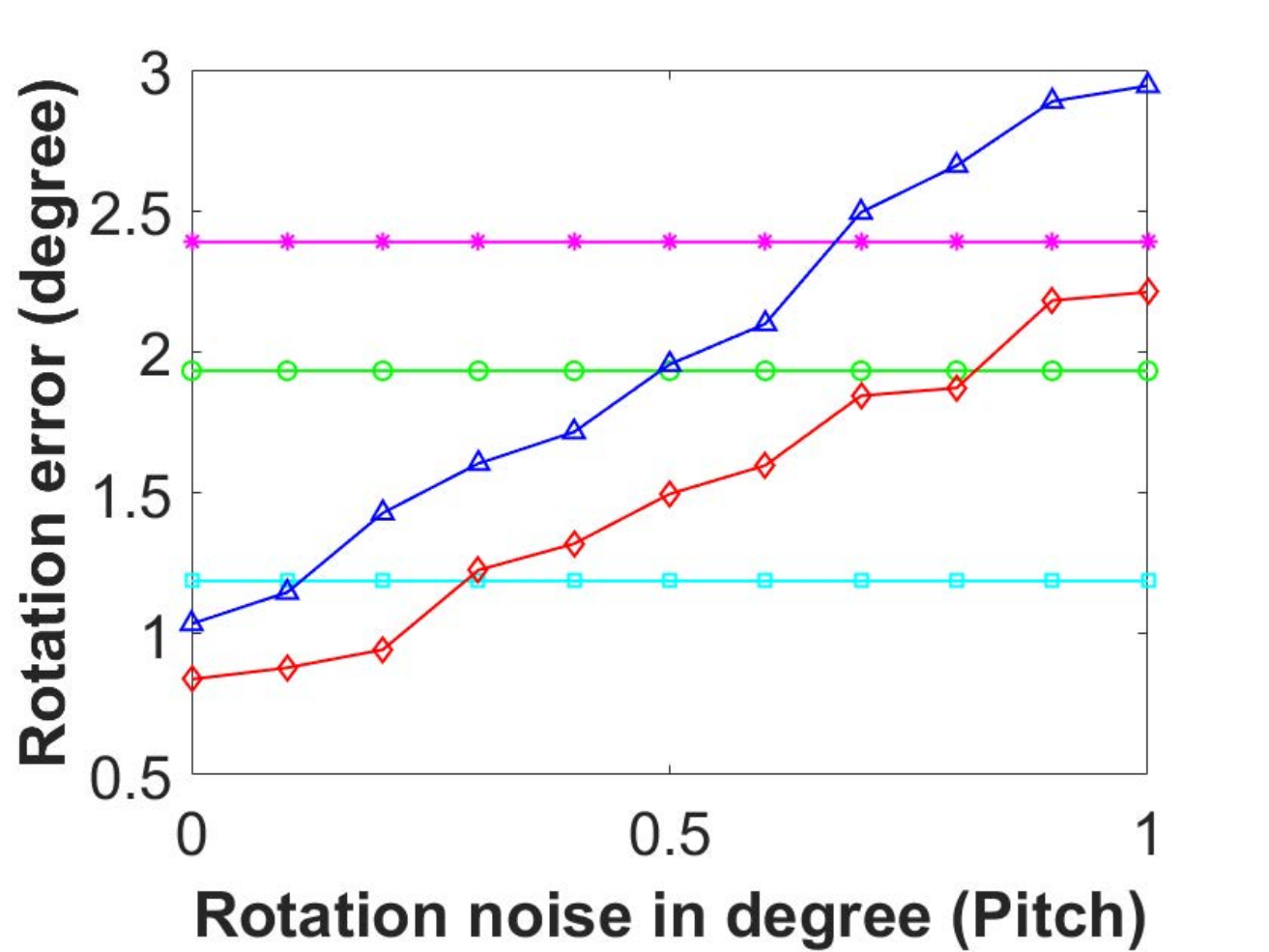}
     \label{fig.IMU angle noise b}
      }
   \subfloat[${\varepsilon _{\bf{t}}}$ - IMU pitch]{
     \centering
     \includegraphics[width=0.45\linewidth]{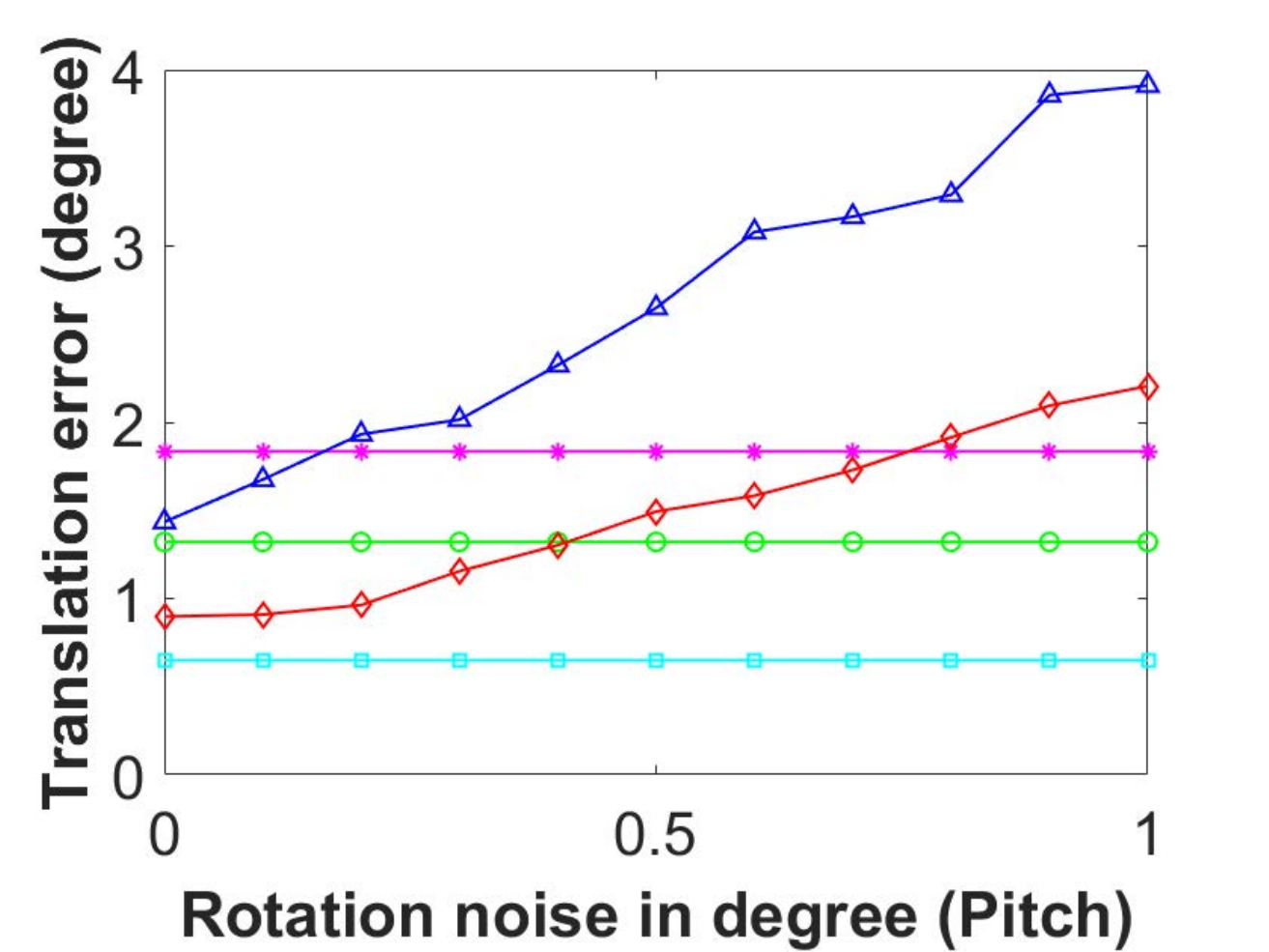}
     \label{fig.IMU angle noise c}
      }
            \vspace{-10pt}
     \subfloat[${\varepsilon _{\bf{R}}}$ - IMU roll]{
     \centering
     \includegraphics[width=0.45\linewidth]{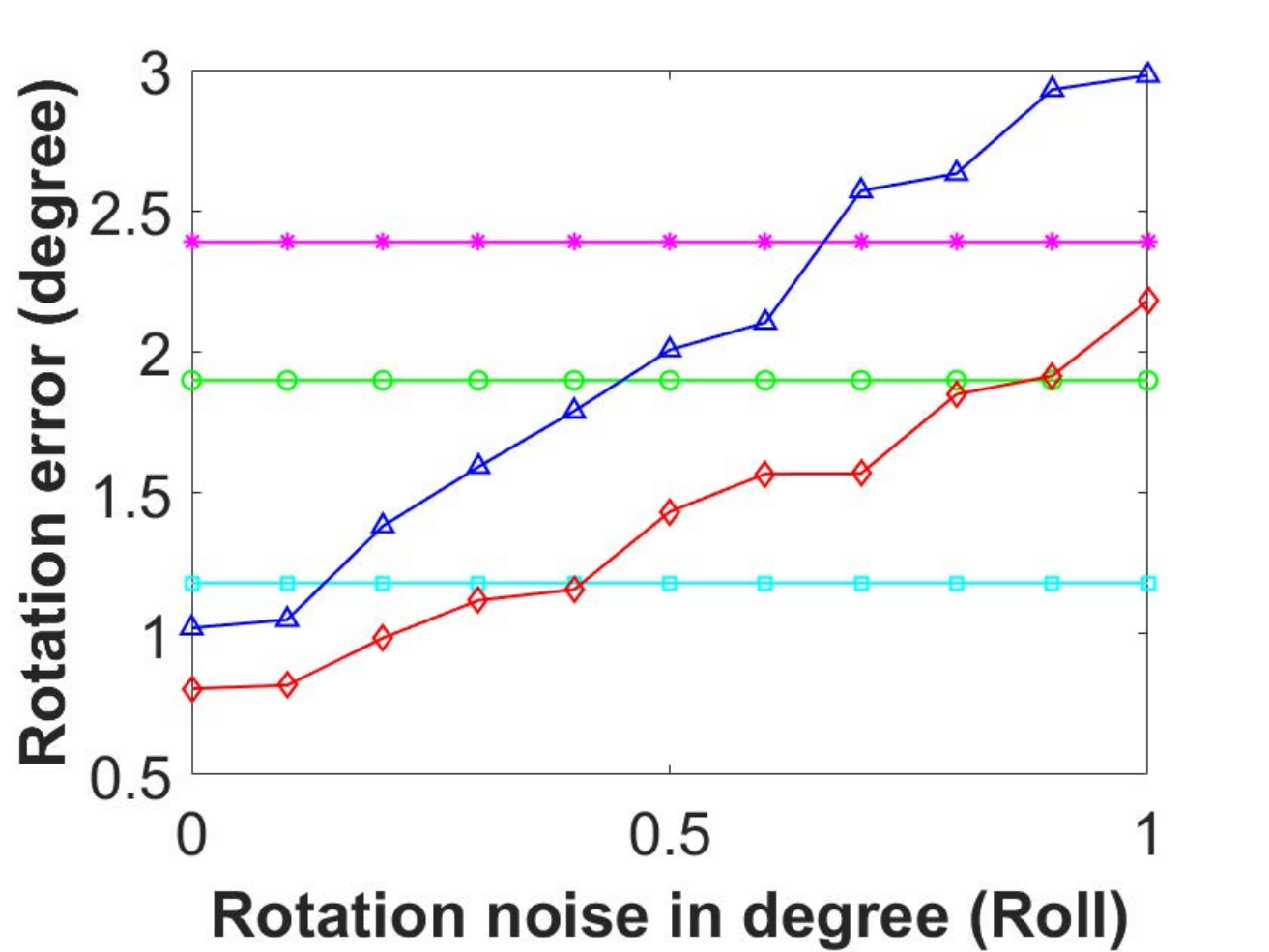}
     \label{fig.IMU angle noise d}
      }
   \subfloat[${\varepsilon _{\bf{t}}}$  - IMU roll]{
     \centering
     \includegraphics[width=0.45\linewidth]{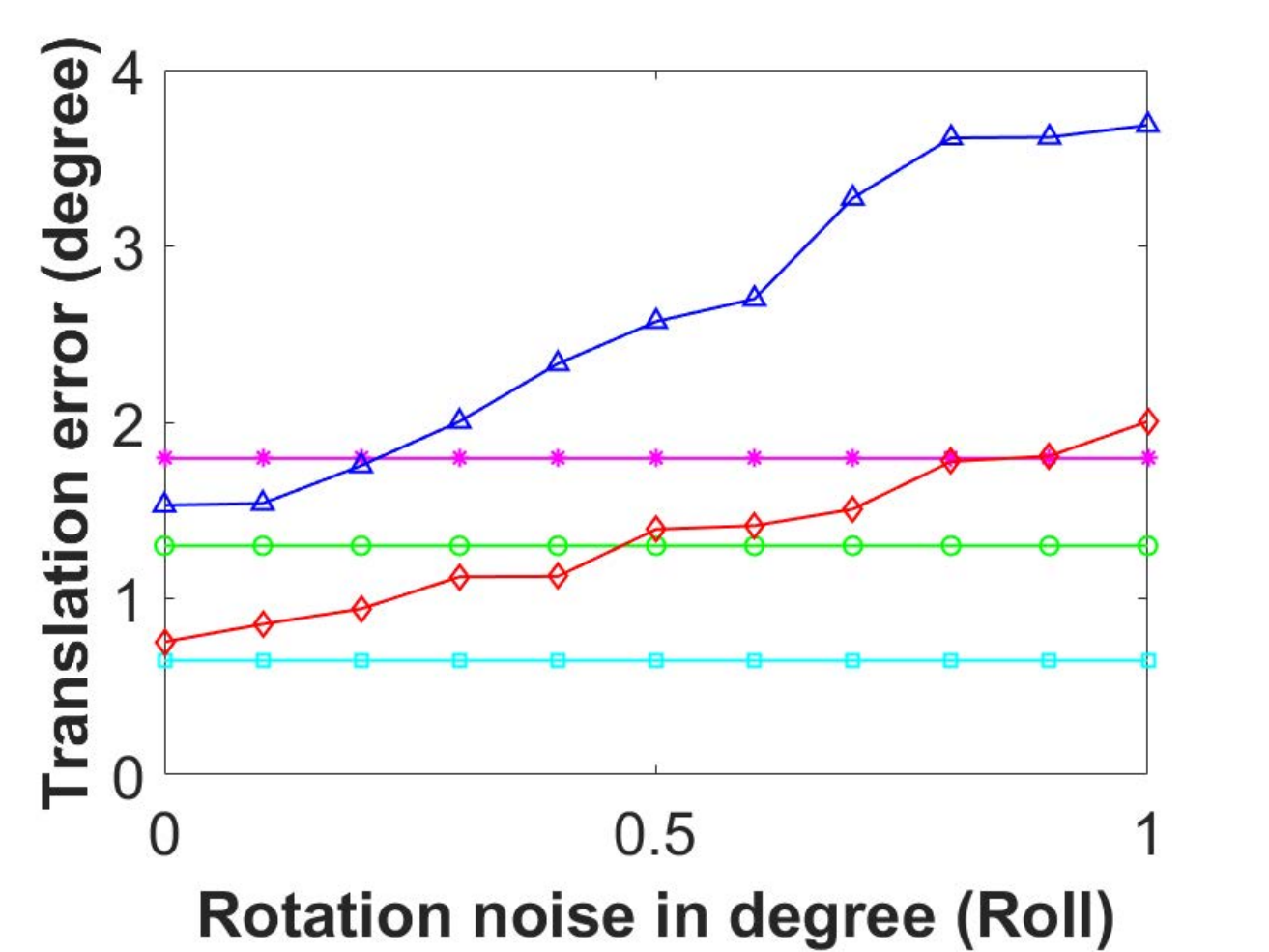}
     \label{fig.IMU angle noise e}
      }
          \centering
    \caption{Methods accuracy with IMU angle noise. (a) Rotation estimation errors with IMU pitch angel noise. (b) Translation estimation errors with IMU pitch angel noise. (c) Rotation estimation errors with IMU roll angel noise. (d) Translation estimation errors with IMU roll angel noise.}

    \label{fig.IMU angle noise1}
        \vspace{-5pt}
\end{figure}

In most cases, a low-cost IMU has an angle accuracy of 0.5°, while that of a high-performance IMU is only 0.02°~\cite{kukelova2010closed}. In practical applications, the accuracy of the IMUs used in cars or modern smartphones is approximately 0.06°~\cite{Ding2021Globally}. To study the effects of IMU angle noise on the algorithm, in the simulation, we set the IMU angle noise ranging from 0° to 1°. The image noise was kept constant at 1 pixel. Since the \verb+8pt-Hartley-2view+, \verb+5pt-Nister-2view+ and \verb+7pt-Hartley-3view+ methods don't use IMUs measurement information, these methods are not affected by the IMU angle noise, so we take the mean values as the final results in Fig. \ref{fig.IMU angle noise1}. 
Although the \verb+4pt-Our-3view+ and \verb+3pt-Our-3view+ methods are affected by IMU angle noise, the rotation accuracy is still superior to the \verb+7pt-Hartley-3view+ method, or even better than the \verb+8pt-Hartley-2view+ method when the inertial navigation error is less than 0.5°. The translation accuracy is more susceptible to the influence of inertial navigation angle noise, but they are still within a reasonable range, and the \verb+3pt-Our-3view+ method is still superior to the \verb+7pt-Hartley-3view+ or \verb+8pt-Hartley-2view+ method.  

\subsection{Experiments on Real-world Data}
The real-world performances of the proposed method were evaluated on the KITTI benchmarks, currently the largest dataset for evaluating computer vision algorithms in autonomous driving scenarios. The KITTI dataset has many sequences, from which sequences 00-10 can provide ground truth values measured by the built-in GPS/IMU units. To evaluate the efficacy of the proposed method in the real world, we compared its performance on all eleven sequences. 
\begin{figure*}[tbp]
  \centering
     \subfloat[View 1]{
     \centering
     \includegraphics[width=0.31\linewidth]{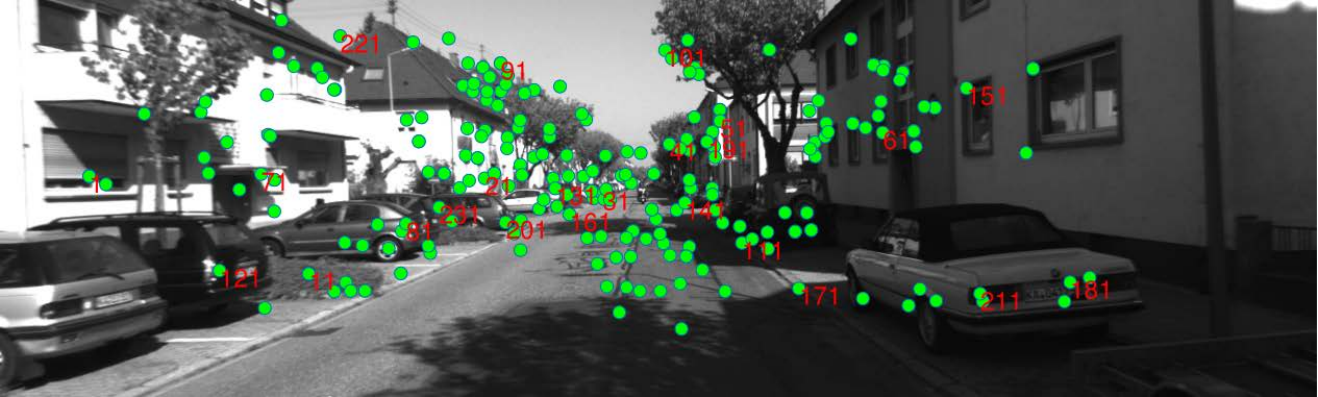}
     \label{fig.MATCH1}
      }
      \subfloat[View 2]{
     \centering
     \includegraphics[width=0.31\linewidth]{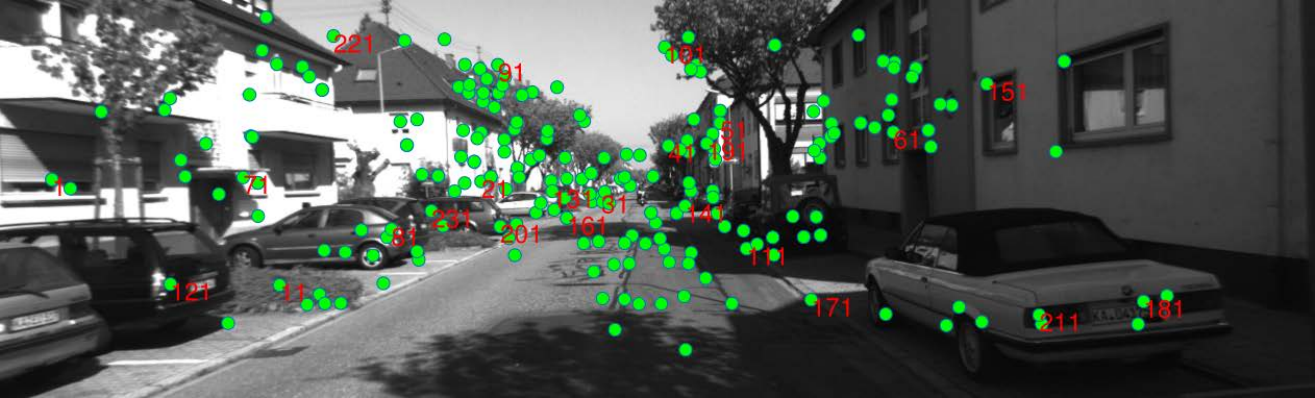}
     \label{fig.MATCH2}
      }
   \subfloat[View 3]{
     \centering
     \includegraphics[width=0.31\linewidth]{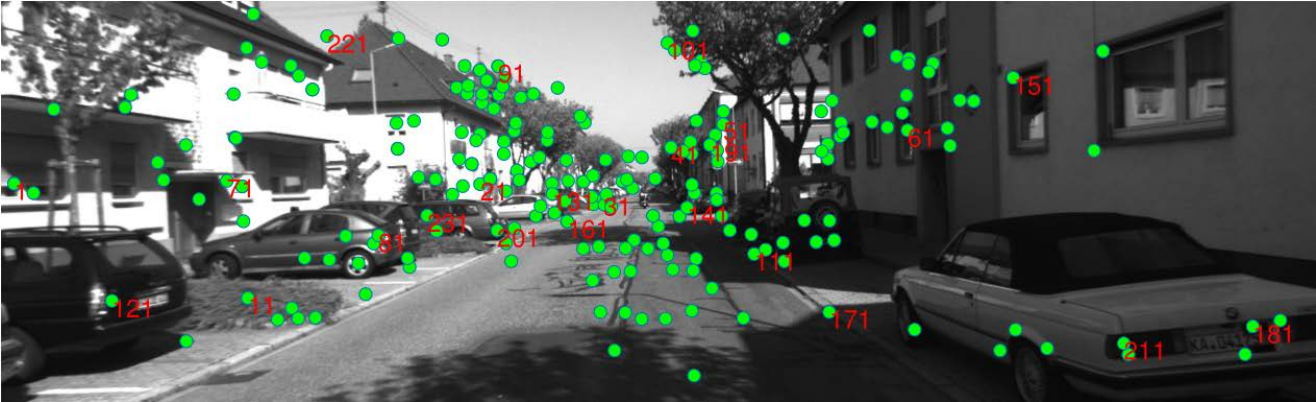}
     \label{fig.MATCH3}
      }
          \centering
    \caption{Point correspondences of triple images. The pictures were taken from sequential images 56 to 58 in sequence 00 of the KITTI dataset.}
    \label{fig.point correspondences}
    \vspace{-10pt}
\end{figure*}

\begin{figure*}[htbp]       
     \subfloat[8pt-Hartley-2view]{
     \centering
     \includegraphics[width=0.18\linewidth]{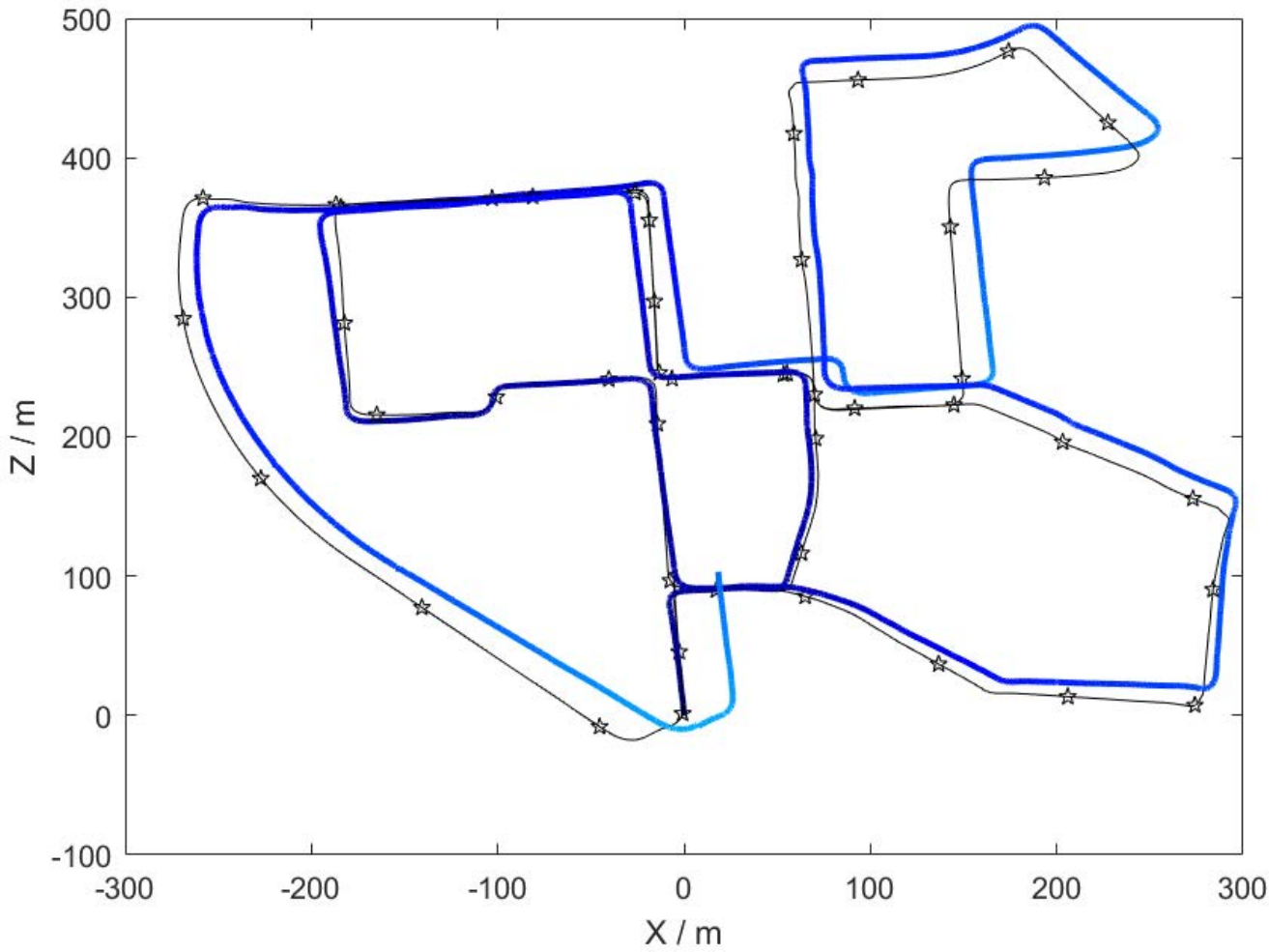}
     \label{fig.guiji a}
      }
   \subfloat[5pt-Nister-2view]{
     \centering
     \includegraphics[width=0.18\linewidth]{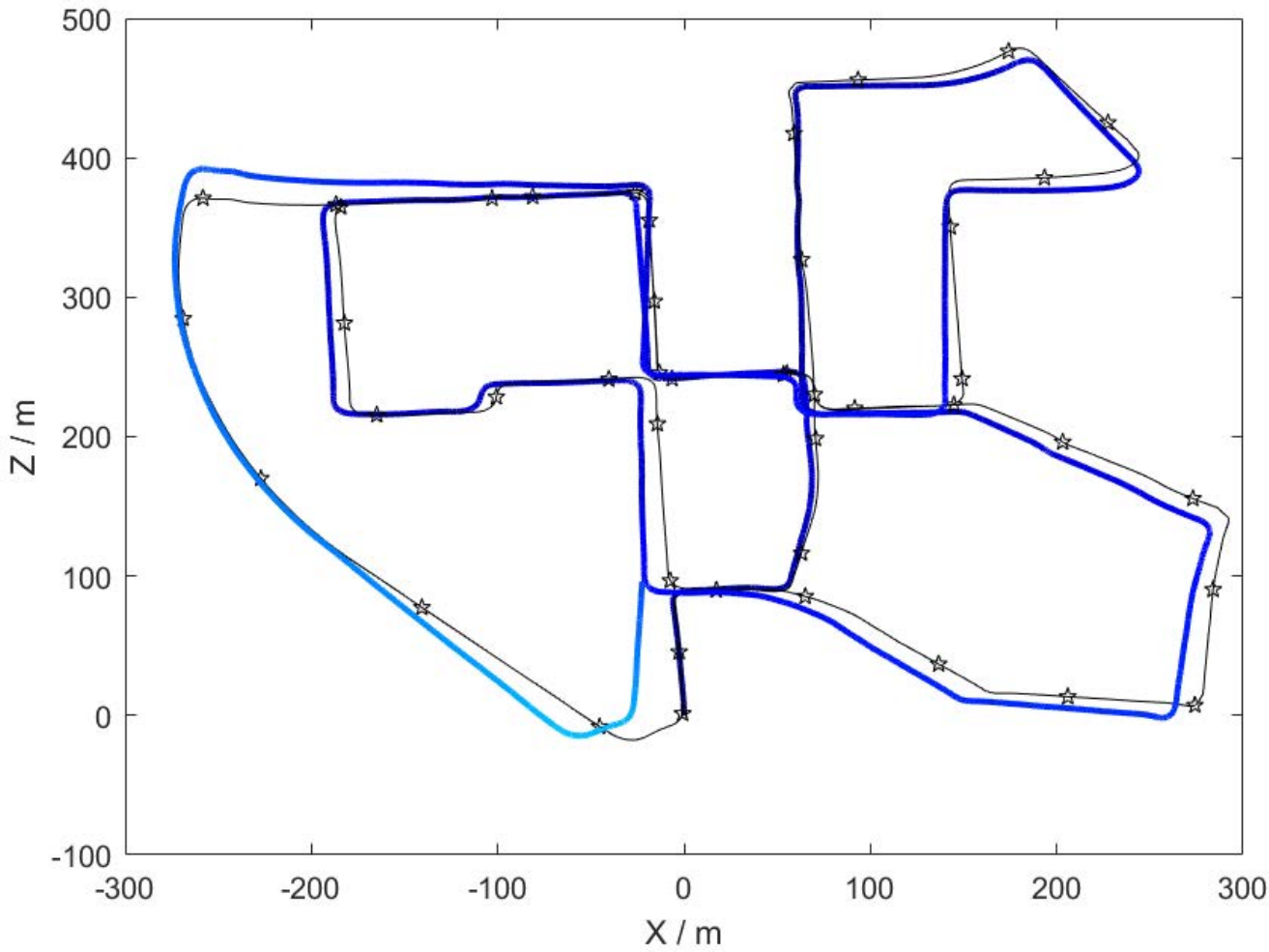}
     \label{fig.guiji b}
      }
     \subfloat[7pt-Hartley-3view]{
     \centering
     \includegraphics[width=0.18\linewidth]{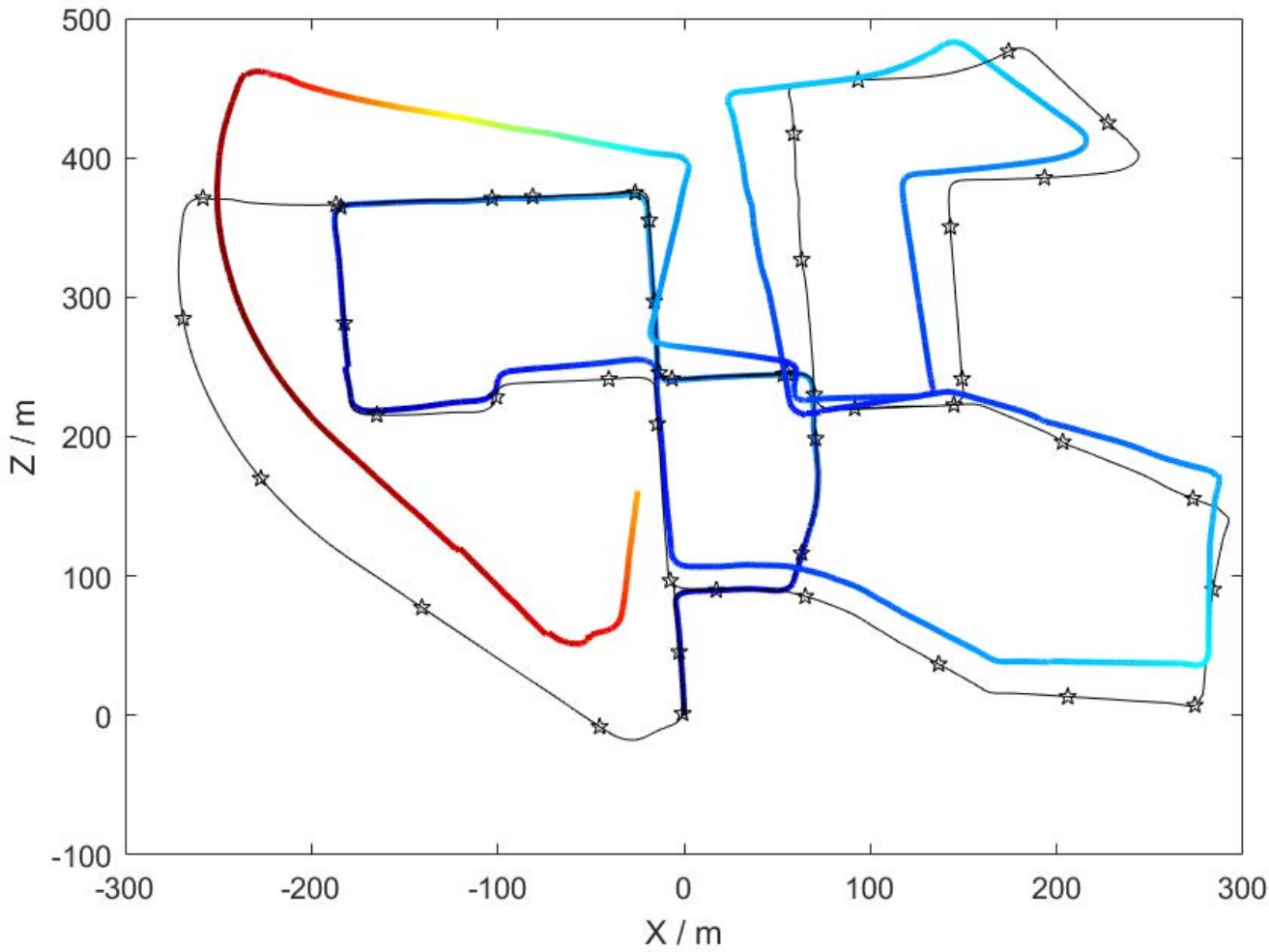}
     \label{fig.guiji c}
      }
   \subfloat[3pt-Our-3view]{
     \centering
     \includegraphics[width=0.18\linewidth]{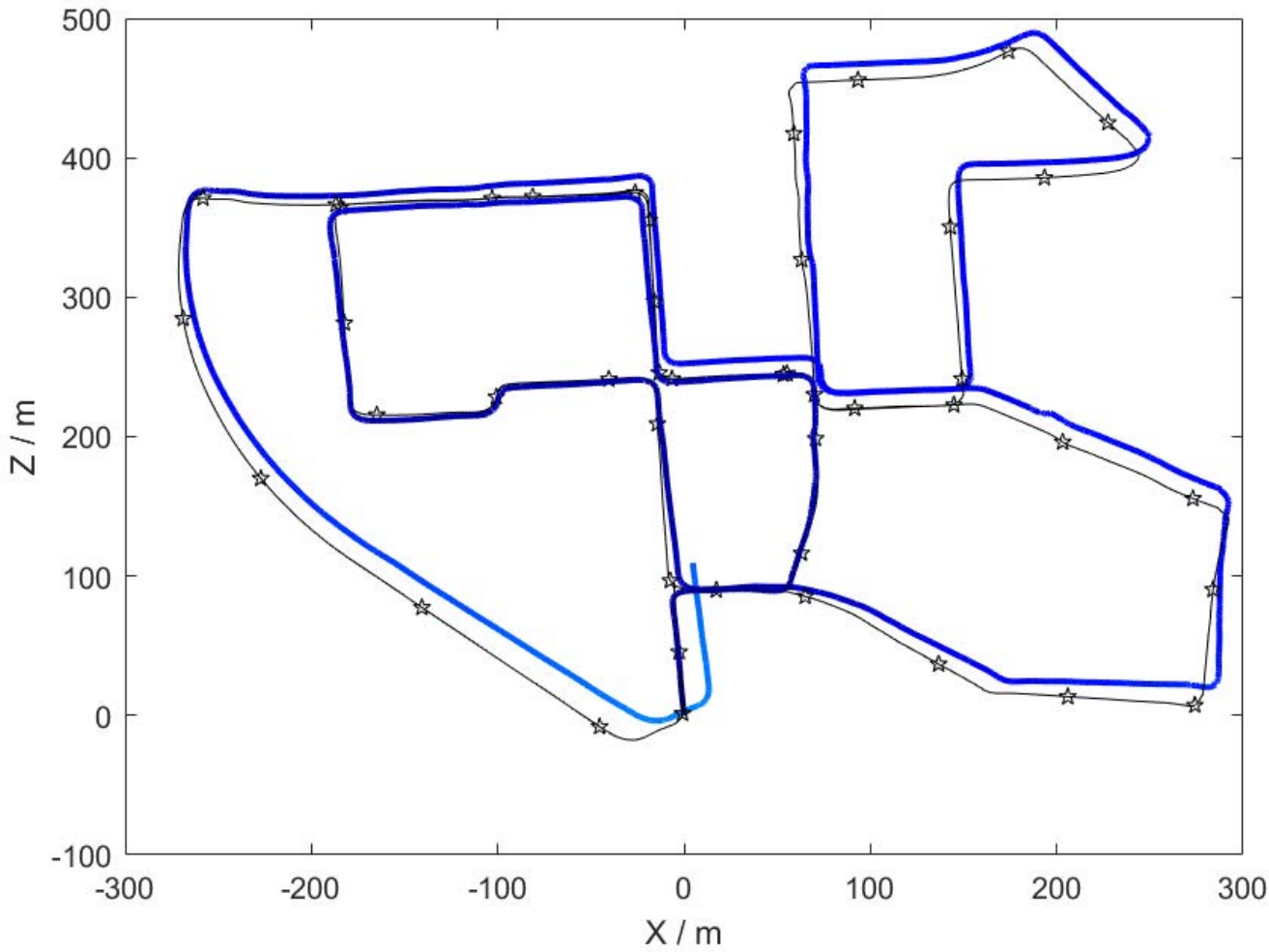}
     \label{fig.guiji d}
      }
      \centering
    \subfloat[4pt-Our-3view]{
     \centering
     \includegraphics[width=0.21\linewidth]{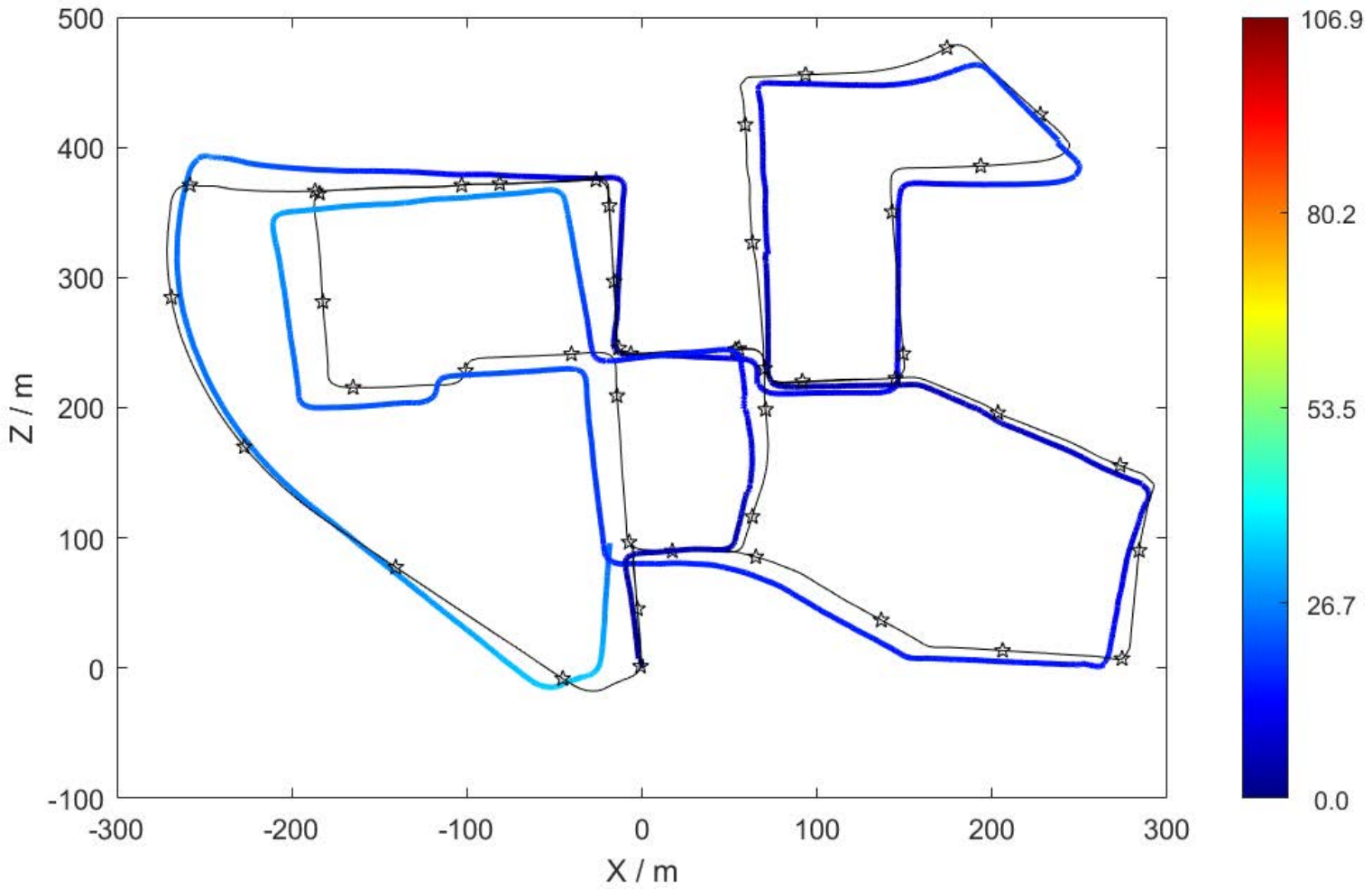}
     \label{fig.guiji e}
      }
    
    \caption{ Estimated visual odometry trajectories for KITTI sequences 00 (unit: meter). (a) Trajectory estimation with 8pt-Hartley-2view method, (b) Trajectory estimation with 5pt-Nister-2view method, (c) Trajectory estimation with 7pt-Hartley-3view method, (d) Trajectory estimation with 3pt-Our-3view method, (e) Trajectory estimation with 4pt-Our-3view method.}

    \label{fig.GUIJI}
     \vspace{-10pt}
\end{figure*}
All methods adopt the SIFT algorithm~\cite{lowe2004distinctive} to extract the point correspondences of triple images. From Fig. \ref{fig.point correspondences}, we can see that there are three consecutive images, and 376 point correspondences (indicated by green dots) can be obtained, and their corresponding relationships are indicated by red numbers. For the sake of visibility, only a few of the red numbers are indicated in the figure.

To eliminate the influence of outliers, all methods are integrated with the RANSAC framework. As for RANSAC parameters, distances are calculated using the Sampson error, and all methods use the same distance threshold. It should be noted that all final pose results are the initial solutions with the largest number of inlier points, that is, the results are not optimized by subsequent optimization algorithms. 
The median errors for rotation estimation and translation estimation are utilized to assess the performance of each method, its detailed results can be seen in Table \ref{tab:realR} and Table \ref{tab:realT}.
 \begin{table}[htbp]
 \vspace{-5pt}
    \caption{ Median rotation errors for KITTI sequences(unit: degree)}
    \label{tab:realR}
        \setlength\tabcolsep{4pt}
    \begin{center}
     \vspace{-10pt}
\begin{tabular}{|c|c|c|c|c|c|}
\hline
Seq. & \begin{tabular}[c]{@{}c@{}}8pt-Hartley\\ -2view\end{tabular} & \begin{tabular}[c]{@{}c@{}}5pt-Nister\\ -2view\end{tabular} & \begin{tabular}[c]{@{}c@{}}7pt-Hartley\\ -3view\end{tabular} & \begin{tabular}[c]{@{}c@{}}3pt-Our\\ -3view\end{tabular} & \begin{tabular}[c]{@{}c@{}}4pt-Our\\ -3view\end{tabular} \\ \hline
00   & 0.160                                                         & 0.086                                                       & 0.118                                                        & \textbf{0.042}                                           & 0.043                                                    \\ \hline
01   & 0.502                                                        & 0.080                                                        & 0.450                                                         & \textbf{0.047}                                           & 0.055                                                    \\ \hline
02   & 0.157                                                        & 0.086                                                       & 0.123                                                        & \textbf{0.042}                                           & \textbf{0.042}                                           \\ \hline
03   & 0.163                                                        & 0.076                                                       & 0.134                                                        & \textbf{0.050}                                            & \textbf{0.050}                                            \\ \hline
04   & 0.172                                                        & 0.074                                                       & 0.135                                                        & \textbf{0.028}                                           & 0.036                                                    \\ \hline
05   & 0.151                                                        & 0.066                                                       & 0.103                                                        & 0.032                                                    & \textbf{0.031}                                           \\ \hline
06   & 0.150                                                         & 0.065                                                       & 0.116                                                        & \textbf{0.031}                                           & 0.034                                                    \\ \hline
07   & 0.135                                                        & 0.069                                                       & 0.107                                                        & \textbf{0.033}                                           & 0.034                                                    \\ \hline
08   & 0.146                                                        & 0.075                                                       & 0.109                                                        & \textbf{0.032}                                           & 0.034                                                    \\ \hline
09   & 0.164                                                        & 0.082                                                       & 0.130                                                         & \textbf{0.038}                                           & 0.039                                                    \\ \hline
10   & 0.154                                                        & 0.081                                                       & 0.120                                                         & \textbf{0.034}                                           & \textbf{0.034}                                           \\ \hline
\end{tabular}
    \end{center}
    \vspace{-15pt}
 \end{table}
 \subsubsection{Rotation errors}
 From Table \ref{tab:realR}, we can see that the \verb+3pt-Our-3view+ method has the smallest rotational errors in the KITTI dataset, followed by the \verb+4pt-Our-3view+ method, which is also consistent with the results of synthetic experiments. The \verb+7pt-Hartley-3view+ and \verb+8pt-Hartley-2view+ methods have the worst rotation accuracy. Overall, the performance of the \verb+3pt-Our-3view+ and \verb+4pt-Our-3view+ methods is 2-3 times more accurate than other methods.
 \subsubsection{Translation errors}
Compared with the rotation errors, the translation errors of each method are bigger. The translation errors have more outliers, which has a certain impact on the accuracy of all the methods. From Table \ref{tab:realT}, we can see that the \verb+5pt-Nister-2view+ method performs best. Similar to the synthetic experiments, the accuracy of the 3pt-Our-3view method performs almost close to the \verb+5pt-Nister-2view+, and even higher in sequences 04, 06, and 07.
\begin{table}[htbp]
 \vspace{-5pt}
    \caption{ Median translation errors for KITTI sequences(unit: degree)}
    \label{tab:realT}
        \setlength\tabcolsep{4pt}
    \begin{center}
     \vspace{-10pt}
\begin{tabular}{|c|c|c|c|c|c|}
\hline
Seq. & \begin{tabular}[c]{@{}c@{}}8pt-Hartley\\ -2view\end{tabular} & \begin{tabular}[c]{@{}c@{}}5pt-Nister\\ -2view\end{tabular} & \begin{tabular}[c]{@{}c@{}}7pt-Hartley\\ -3view\end{tabular} & \begin{tabular}[c]{@{}c@{}}3pt-Our\\ -3view\end{tabular} & \begin{tabular}[c]{@{}c@{}}4pt-Our\\ -3view\end{tabular} \\ \hline
00   & 1.498                                                        & \textbf{1.178}                                              & 1.296                                                        & 1.296                                                    & 1.436                                                    \\ \hline
01   & 3.148                                                        & \textbf{2.095}                                              & 3.874                                                        & 2.435                                                    & 3.53                                                     \\ \hline
02   & 1.411                                                        & \textbf{1.124}                                              & 1.240                                                         & 1.238                                                    & 1.381                                                    \\ \hline
03   & 1.780                                                         & \textbf{1.306}                                              & 1.663                                                        & 1.587                                                    & 1.863                                                    \\ \hline
04   & 1.286                                                        & 1.076                                                       & 1.059                                                        & \textbf{0.789}                                           & 1.221                                                    \\ \hline
05   & 1.294                                                        & \textbf{0.970}                                               & 1.077                                                        & 1.030                                                     & 1.217                                                    \\ \hline
06   & 1.040                                                         & 0.812                                                       & 0.900                                                          & \textbf{0.782}                                           & 0.985                                                    \\ \hline
07   & 1.692                                                        & 1.300                                                         & 1.490                                                         & \textbf{1.267}                                           & 1.562                                                    \\ \hline
08   & 1.653                                                        & \textbf{1.368}                                              & 1.499                                                        & 1.429                                                    & 1.625                                                    \\ \hline
09   & 1.393                                                        & \textbf{1.008}                                              & 1.217                                                        & 1.034                                                    & 1.237                                                    \\ \hline
10   & 1.609                                                        & \textbf{1.058}                                              & 1.336                                                        & 1.256                                                    & 1.367                                          \\ \hline
\end{tabular}
    \end{center}
    \vspace{-10pt}
 \end{table}
 
To visualize the comparison of all methods, we also plotted the camera's trajectory in Fig. \ref{fig.GUIJI}. Since the estimated translation of the monocular camera is up to scale, all the methods use the ground truth scale to recover its real translation. In Fig. \ref{fig.GUIJI}, the black trajectory represents the ground truth, the colored trajectory represents the estimated values, and its color encodes the magnitude of the absolute trajectory error (ATE)~\cite{sturm2012benchmark}. Because of space constraints, only the trajectory of sequence 00 is presented. As can be seen from Fig. \ref{fig.GUIJI}, compared with the \verb+7pt-Hartley-3view+ method, the \verb+3pt-Our-3view+ and \verb+4pt-Our-3view+ methods have smaller ATE, which also indirectly illustrates the accuracy of our proposed methods.

\section{CONCLUSION}

In this paper, we derived the methods based on 4 or 3 point correspondences to solve the 3-view problem with IMU measurement data. First, by using the angle information measured by IMUs, the 3-view problem can be effectively simplified. Then, by utilizing either 4 point correspondences or 3 point correspondences, we can derive the linear closure solution and the minimal solution, respectively. Experiments on synthetic data and real-world data show that our methods are accurate and feasible. In the application of autonomous driving or drone navigation, our algorithm can provide an effective backup option when 2-view methods fail.

\bibliographystyle{IEEEtran}
\bibliography{references}

\begin{thebibliography}{10}
\providecommand{\url}[1]{#1}
\csname url@rmstyle\endcsname
\providecommand{\newblock}{\relax}
\providecommand{\bibinfo}[2]{#2}
\providecommand\BIBentrySTDinterwordspacing{\spaceskip=0pt\relax}
\providecommand\BIBentryALTinterwordstretchfactor{4}
\providecommand\BIBentryALTinterwordspacing{\spaceskip=\fontdimen2\font plus
\BIBentryALTinterwordstretchfactor\fontdimen3\font minus
  \fontdimen4\font\relax}
\providecommand\BIBforeignlanguage[2]{{%
\expandafter\ifx\csname l@#1\endcsname\relax
\typeout{** WARNING: IEEEtran.bst: No hyphenation pattern has been}%
\typeout{** loaded for the language `#1'. Using the pattern for}%
\typeout{** the default language instead.}%
\else
\language=\csname l@#1\endcsname
\fi
#2}}

\bibitem{li20134}
B.~Li, L.~Heng, G.~H. Lee, and M.~Pollefeys, ``A 4-point algorithm for relative
  pose estimation of a calibrated camera with a known relative rotation
  angle,'' in \emph{IEEE/RSJ International Conference on Intelligent Robots and
  Systems}, 2013, pp. 1595--1601.

\bibitem{campos2021orb}
C.~Campos, R.~Elvira, J.~J.~G. Rodr{\'\i}guez, J.~M. Montiel, and J.~D.
  Tard{\'o}s, ``Orb-slam3: An accurate open-source library for visual,
  visual-inertial, and multimap slam,'' \emph{IEEE Transactions on Robotics},
  vol.~37, no.~6, pp. 1874--1890, 2021.

\bibitem{mur2017orb}
R.~Mur-Artal and J.~D. Tard{\'o}s, ``Orb-slam2: An open-source slam system for
  monocular, stereo, and rgb-d cameras,'' \emph{IEEE Transactions on Robotics},
  vol.~33, no.~5, pp. 1255--1262, 2017.

\bibitem{yu2023globally}
Z.~Yu, B.~Guan, S.~Liang, Z.~Li, S.~Ye, and Q.~Yu, ``Globally optimal relative
  pose estimation using affine correspondences with known vertical direction,''
  \emph{IEEE Transactions on Instrumentation and Measurement}, vol.~72, pp.
  1--12, 2023.

\bibitem{nister2006four}
D.~Nist{\'e}r and F.~Schaffalitzky, ``Four points in two or three calibrated
  views: Theory and practice,'' \emph{International Journal of Computer
  Vision}, vol.~67, pp. 211--231, 2006.

\bibitem{ding2023minimal}
Y.~Ding, C.-H. Chien, V.~Larsson, K.~{\AA}str{\"o}m, and B.~Kimia, ``Minimal
  solutions to generalized three-view relative pose problem,'' in \emph{IEEE
  International Conference on Computer Vision}, 2023, pp. 8156--8164.

\bibitem{julia2018critical}
L.~F. Juli{\`a} and P.~Monasse, ``A critical review of the trifocal tensor
  estimation,'' in \emph{Pacific-Rim Symposium on Image and Video Technology},
  2018, pp. 337--349.

\bibitem{fischler1981random}
M.~A. Fischler and R.~C. Bolles, ``Random sample consensus: a paradigm for
  model fitting with applications to image analysis and automated
  cartography,'' \emph{Communications of the ACM}, vol.~24, no.~6, pp.
  381--395, 1981.

\bibitem{nister2004efficient}
D.~Nist{\'e}r, ``An efficient solution to the five-point relative pose
  problem,'' \emph{Transactions on Pattern Analysis and Machine Intelligence},
  vol.~26, no.~6, pp. 756--770, 2004.

\bibitem{hartley2003multiple}
R.~Hartley and A.~Zisserman, \emph{Multiple view geometry in computer
  vision}.\hskip 1em plus 0.5em minus 0.4em\relax Cambridge university press,
  2003.

\bibitem{holt1994motion}
R.~J. Holt and A.~N. Netravali, ``Motion and structure from line
  correspondences: Some further results,'' \emph{International Journal of
  Imaging Systems and Technology}, vol.~5, no.~1, pp. 52--61, 1994.

\bibitem{fabbri2020trplp}
R.~Fabbri, T.~Duff, H.~Fan, M.~H. Regan, D.~d. C.~d. Pinho, E.~Tsigaridas,
  C.~W. Wampler, J.~D. Hauenstein, P.~J. Giblin, B.~Kimia, \emph{et~al.},
  ``Trplp-trifocal relative pose from lines at points,'' in \emph{IEEE
  Conference on Computer Vision and Pattern Recognition}, 2020, pp.
  12\,073--12\,083.

\bibitem{guan2022trifocal}
B.~Guan, P.~Vasseur, and C.~Demonceaux, ``Trifocal tensor and relative pose
  estimation from 8 lines and known vertical direction,'' in \emph{IEEE/RSJ
  International Conference on Intelligent Robots and Systems}, 2022, pp.
  6001--6008.

\bibitem{heyden1997reconstruction}
A.~Heyden, ``Reconstruction from image sequences by means of relative depths,''
  \emph{International Journal of Computer Vision}, vol.~24, no.~2, pp.
  155--161, 1997.

\bibitem{ressl2002minimal}
C.~Ressl, ``A minimal set of constraints and a minimal parameterization for the
  trifocal tensor,'' \emph{International Archives of Photogrammetry Remote
  Sensing and Spatial Information Sciences}, vol.~34, no. 3/A, pp. 277--282,
  2002.

\bibitem{nordberg2009minimal}
K.~Nordberg, ``A minimal parameterization of the trifocal tensor,'' in
  \emph{IEEE Conference on Computer Vision and Pattern Recognition}, 2009, pp.
  1224--1230.

\bibitem{faugeras1998nonlinear}
O.~Faugeras and T.~Papadopoulo, ``A nonlinear method for estimating the
  projective geometry of 3 views,'' in \emph{IEEE International Conference on
  Computer Vision}, 1998, pp. 477--484.

\bibitem{ponce2014trinocular}
J.~Ponce and M.~Hebert, ``Trinocular geometry revisited,'' in \emph{IEEE
  Conference on Computer Vision and Pattern Recognition}, 2014, pp. 17--24.

\bibitem{hruby2022learning}
P.~Hruby, T.~Duff, A.~Leykin, and T.~Pajdla, ``Learning to solve hard minimal
  problems,'' in \emph{IEEE Conference on Computer Vision and Pattern
  Recognition}, 2022, pp. 5532--5542.

\bibitem{xu2024accurate}
Z.~Xu, Y.~He, H.~Wei, B.~Xu, B.~Xie, and Y.~Wu, ``An accurate and real-time
  relative pose estimation from triple point-line images by decoupling rotation
  and translation,'' \emph{arXiv preprint arXiv:2403.11639}, 2024.

\bibitem{kukelova2008automatic}
Z.~Kukelova, M.~Bujnak, and T.~Pajdla, ``Automatic generator of minimal problem
  solvers,'' in \emph{European Conference on Computer Vision}, 2008, pp.
  302--315.

\bibitem{larsson2018beyond}
V.~Larsson, M.~Oskarsson, K.~Astrom, A.~Wallis, Z.~Kukelova, and T.~Pajdla,
  ``Beyond grobner bases: Basis selection for minimal solvers,'' in \emph{IEEE
  Conference on Computer Vision and Pattern Recognition}, 2018, pp. 3945--3954.

\bibitem{larsson2017efficient}
V.~Larsson, K.~Astrom, and M.~Oskarsson, ``Efficient solvers for minimal
  problems by syzygy-based reduction,'' in \emph{IEEE Conference on Computer
  Vision and Pattern Recognition}, 2017, pp. 820--829.

\bibitem{li2020gaps}
B.~Li and V.~Larsson, ``Gaps: Generator for automatic polynomial solvers,''
  \emph{arXiv preprint arXiv:2004.11765}, 2020.

\bibitem{bhayani2020sparse}
S.~Bhayani, Z.~Kukelova, and J.~Heikkila, ``A sparse resultant based method for
  efficient minimal solvers,'' in \emph{IEEE Conference on Computer Vision and
  Pattern Recognition}, 2020, pp. 1770--1779.

\bibitem{martyushev2023automatic}
E.~Martyushev, S.~Bhayani, and T.~Pajdla, ``Automatic solver generator for
  systems of laurent polynomial equations,'' \emph{arXiv preprint
  arXiv:2307.00320}, 2023.

\bibitem{kukelova2017clever}
Z.~Kukelova, J.~Kileel, B.~Sturmfels, and T.~Pajdla, ``A clever elimination
  strategy for efficient minimal solvers,'' in \emph{IEEE Conference on
  Computer Vision and Pattern Recognition}, 2017, pp. 4912--4921.

\bibitem{ding2021general}
Y.~Ding, Y.~Su, C.~Xu, J.~Yang, and H.~Kong, ``A general elimination strategy
  for camera motion estimation,'' in \emph{2021 IEEE International Conference
  on Robotics and Automation}.\hskip 1em plus 0.5em minus 0.4em\relax IEEE,
  2021, pp. 9333--9339.

\bibitem{sommese2005numerical}
A.~J. Sommese, C.~W. Wampler, \emph{et~al.}, \emph{The Numerical solution of
  systems of polynomials arising in engineering and science}.\hskip 1em plus
  0.5em minus 0.4em\relax World Scientific, 2005.

\bibitem{chien2022gpu}
C.-H. Chien, H.~Fan, A.~Abdelfattah, E.~Tsigaridas, S.~Tomov, and B.~Kimia,
  ``Gpu-based homotopy continuation for minimal problems in computer vision,''
  in \emph{IEEE Conference on Computer Vision and Pattern Recognition}, 2022,
  pp. 15\,765--15\,776.

\bibitem{guan2023minimal}
B.~Guan, J.~Zhao, D.~Barath, and F.~Fraundorfer, ``Minimal solvers for relative
  pose estimation of multi-camera systems using affine correspondences,''
  \emph{International Journal of Computer Vision}, vol. 131, no.~1, pp.
  324--345, 2023.

\bibitem{kukelova2010closed}
Z.~Kukelova, M.~Bujnak, and T.~Pajdla, ``Closed-form solutions to minimal
  absolute pose problems with known vertical direction,'' in \emph{Asian
  Conference on Computer Vision}, 2010, pp. 216--229.

\bibitem{Ding2021Globally}
Y.~Ding, D.~Barath, J.~Yang, H.~Kong, and Z.~Kukelova, ``Globally optimal
  relative pose estimation with gravity prior,'' in \emph{IEEE Conference on
  Computer Vision and Pattern Recognition}, 2021, pp. 394--403.

\bibitem{lowe2004distinctive}
D.~G. Lowe, ``Distinctive image features from scale-invariant keypoints,''
  \emph{International journal of computer vision}, vol.~60, pp. 91--110, 2004.

\bibitem{sturm2012benchmark}
J.~Sturm, N.~Engelhard, F.~Endres, W.~Burgard, and D.~Cremers, ``A benchmark
  for the evaluation of rgb-d slam systems,'' in \emph{IEEE/RSJ International
  Conference on Intelligent Robots and Systems}, 2012, pp. 573--580.

\end{thebibliography}

\end{document}